\documentclass{article}
\usepackage{kr}

\usepackage{times}
\usepackage{soul}
\usepackage{url}
\usepackage[hidelinks]{hyperref}
\usepackage[utf8]{inputenc}
\usepackage[small]{caption}
\usepackage{graphicx}
\usepackage{amsmath}
\usepackage{amssymb}
\usepackage{amsthm}
\usepackage{booktabs}
\usepackage{algorithm}
\usepackage{algorithmic}
\usepackage{wasysym}
\usepackage{enumerate}
\usepackage{fvextra}
\usepackage{relsize}
\fvset{fontfamily=courier,fontsize=\relsize{-1},frame=none,numbers=none,numbersep=4pt,commentchar=!,commandchars=\\\{\}}
\newcommand\atom{\Verb*[fontfamily=courier,fontsize=\relsize{-1}]}

\newtheorem{definition}{Definition}
\newtheorem{theorem}[definition]{Theorem}

\title{Learning First-Order Representations for Planning \\ from Black-Box States: New Results}
\author{%
  Ivan D. Rodriguez$^1$ \and
  Blai Bonet$^1$ \and
  Javier Romero$^2$ \and
  Hector Geffner$^{3}$
  \affiliations
  \textsuperscript{\rm 1}{Universitat Pompeu Fabra, Spain} \\
  \textsuperscript{\rm 2}{University of Potsdam, Germany} \\
  \textsuperscript{\rm 3}{ICREA and Universitat Pompeu Fabra, Spain} \\
  \emails
  \{ivandanielra,bonetblai\}@gmail.com, javier@cs.uni-potsdam.de, hector.geffner@upf.edu
}

\usepackage{booktabs}
\usepackage{xspace}
\usepackage{todonotes}
\presetkeys{todonotes}{inline}{}

\usepackage{pifont}% http://ctan.org/pkg/pifont
\newcommand{\cmark}{\ding{51}}%
\newcommand{\xmark}{\ding{55}}%

\usepackage{csvsimple}

\newcommand{\tup}[1]{\langle #1 \rangle}

\newcommand{\Omit}[1]{}
\newcommand{\pair}[1]{\langle #1 \rangle}

\newcommand{\citeay}[1]{\citeauthor{#1} (\citeyear{#1})}

\newcommand{\alert}[1]{\textcolor{red}{\bf #1}}

\newcommand{\clingo}{\textsc{clingo}\xspace}
\newcommand{\aspg}{\textsc{asp}($G$)\xspace}

\begin{document}
\allowdisplaybreaks

\maketitle

\begin{abstract}
  Recently \citeauthor{bonet:ecai2020} have shown that first-order representations for
  planning domains can be learned from the structure of the state space
  without any prior knowledge about the action schemas or domain predicates.
  For this, the learning problem is formulated as the search for a simplest first-order domain
  description $D$ that along with information about instances $I_i$ (number of objects
  and initial state) determine state space graphs $G(P_i)$ that match the observed state graphs
  $G_i$ where $P_i = \tup{D,I_i}$. The search is cast and solved approximately by means of
  a SAT solver that is called over a large family of propositional theories that differ
  just in the parameters encoding the possible number of action schemas and domain predicates,
  their arities, and the number of objects. In this work, we push the limits of these
  learners by moving to an answer set programming (ASP) encoding using the \clingo system.
  The new encodings are more transparent and concise, extending the range of possible models
  while facilitating their exploration. We show that the domains introduced by \citeauthor{bonet:ecai2020}
  can be solved more efficiently in the new approach, often optimally, and furthermore, that the approach can be
  easily extended to handle partial information about the state graphs as well as noise that prevents some
  states from being distinguished.
\end{abstract}

\section{Introduction}

One of the main research challenges in AI is how to bring together learning and reasoning,
and in particular, learning approaches that can deal with non-symbolic inputs
and reasoning approaches that require first-order symbolic inputs \cite{josh,marcus1,pearl,darwiche}.
On one hand,  pure data-based approaches like deep learning  produce black boxes
that are hard to understand and which do not generalize well; on the other,
model-based approaches, in particular those that require first-order representations,
require models  which are normally crafted by hand.

A concrete challenge for  making  the best of both data-based learners and model-based reasoners
is to learn from data the type of representations that are required by solvers \cite{geffner:ijcai2018}.
In this work we are particularly interested in learning the first-order symbolic representations that are used
in classical planning in languages such as PDDL \cite{pddl,pddl:book}. These languages have
evolved throughout years of research and  exhibit a number of benefits concerning both
generalization and reusability. Planning problems in PDDL-like  languages are expressed
in two parts: a domain $D$ that expresses the action schemas and their preconditions and effects in terms
of a fixed set of domain predicates, and  instance information $I$ that details the objects (names) and
the ground atoms that are  true in the initial situation and  the goal. The domain $D$ and the instance information $I$,  together,
define a complete  planning instance $P=\tup{D,I}$. There is, however,  an infinite collection of planning instances
that can be defined over the same domain,  and the action schemas and predicates provide a language for  capturing
what is common in all of them. Thus, if one manages to learn the domain predicates and schemas from some instances,
one learns a representation that applies to all other domain instances as well.

Two approaches have been recently proposed for learning first-order planning representations
from non-symbolic data. In one case, the input data corresponds to one or more state graphs
$G_i$ assumed to originate from  hidden planning instances $P_i=\tup{D,I_i}$  that need to be uncovered.
In these input graphs, the nodes  correspond to different states,  and the  edges correspond
to the  possible state transitions. The states are black boxes and nothing is assumed to be known about
their structure  except that states associated with different nodes must be different
\cite{bonet:ecai2020}. In the second case, the input involves state trajectories associated with
an instance with each state represented by an image.  Propositional and first-order action representations
are then obtained through the use of a class of (variational) autoencoders \cite{vae},
where the input images must be recovered in the output  of a deep neural network by going through
a categorical representation \cite{asai:fol}.
The two approaches appeal to different principles for uncovering the representations:
in the first case, the learned representations must recover the input graphs;
in the second, the images associated with the states.  The approach  based on \emph{graph recovery}
yields crisp symbolic representations that match the intended models well;
the second approach based on \emph{image recovery} yields symbolic representations
that are less crisp  but which are  grounded on the images,
and make less assumptions about the  inputs than the graph recovery approach
that requires the set of  trajectories (graphs) to be  complete and noise-free.

In this work, we explore variations and extensions of the approach %to representation learning for planning
proposed by \citeauthor{bonet:ecai2020} that make it more scalable and more robust.
In their work, the  search for the simplest PDDL models $P_i=\tup{D,I_i}$ that account for the input graphs $G_i$
is formulated  and solved, approximately, by means of a SAT solver that is called over a  large family  of propositional theories
differing in the  parameters encoding the  possible number of action schemas and domain predicates,
their arities, and the number of objects. We move from a low-level SAT encoding to a high-level
\emph{answer set programming (ASP)} encoding \cite{brewka:asp,vladimir:asp} 
using the \clingo system \cite{torsten:asp,torsten:clingo}.
The new  encoding is more transparent and  more concise, and extends  the  range of
possible models while  facilitating  their  exploration. We show that the domains introduced by \citeauthor{bonet:ecai2020}
can be  solved \emph{more efficiently} in the new approach,  in many cases optimally, and  furthermore  that  simple extensions suffice
to overcome some of the limitations,   like {the assumption that the input graphs are complete and noise-free.}
Indeed, the new encodings  can handle naturally \emph{partial information about the state graphs},  as
well as \emph{noise that prevents some states from being distinguished from other states.}

The paper is organized as follows. We first review related work and the learning formulation advanced by
\citeauthor{bonet:ecai2020}.
Then we solve this formulation via ASP, consider a number of extensions and optimizations,
and present the experimental results  and the extensions for dealing with incomplete and noisy  samples of the
input graphs.

\section{Related Work}

The paper builds on prior work by \citeay{bonet:ecai2020} and is related to the work by \citeay{asai:fol}, both focused
on the problem of learning first-order symbolic representations of planning domains from non-symbolic data. % \cite{bonet:ecai2020,asai:fol}.
The language of these representations is a subset of PDDL
which is suitable for transferring knowledge learned for some planning instances
to others. This is knowledge about the \emph{general domain models}. Another form
of knowledge that can be transferred among instances of the same domain
is given by \emph{general policies or plans} \cite{srivastava08learning,bonet:icaps2009,hu:generalized,BelleL16}.
A general policy provides a full-detailed strategy for solving
a collection of problems.
%%For example, a general strategy for achieving the goal $on(x,y)$
%%in Blocksworld for two blocks $x$ and $y$ can achieve first
%%$clear(x)$ by removing the blocks above $x$ one by one,
%%then $clear(y)$ in the same way while avoiding placing
%%blocks on $x$, and finally moving $x$ to $y$.
%%The policy must work for any number of blocks with any names,
%%starting from any configuration.\footnote{Actually, this policy is not fully general
%%and does not work when block $x$ is above $y$ initially.}
Approaches for learning such general plans from some instances
have been developed as well \cite{khardon:generalized,martin:generalized,fern:generalized,bonet:aaai2019,guillem:aaai2021},
some of them relying on deep learning techniques \cite{trevizan:dl,sanner:dl,fern:dl,mausam:dl}.
Since these approaches require a first-order representation of the planning domains,
learning these representations provides a necessary step for learning the general policies.

{Deep reinforcement learning} (DRL) methods \cite{atari}
have been used to learn general policies over high-dimensional perceptual spaces
without using or producing symbolic knowledge \cite{sid:sokoban,babyAI,pineau}.
Yet by not constructing first-order representations involving objects and relations,
their ability to generalize appears to be limited.
Recent work in {deep symbolic relational reinforcement learning}
\cite{shanahan:review,shanahan:predinet} attempts to account for objects and relations through the use of suitable neural architectures,
but the gap between the {low-level techniques} used and the {high-level representations} required is large.
More recently, model-based DRL approaches have been shown to learn informative latent representations and have
achieved considerable success in specific settings and video-games, but their abilities for generalization in the presence
of new objects has not been explored \cite{model-based-drl:planet,model-based-drl:muzero,model-based-drl:dreamer2}.

Finally, there is large body of work on learning first-order planning representations given partial knowledge about
the instances and domains \cite{oo-mdp1,yang:model-learning,review-learning-planning,sergio:model-recog,locm}; e.g.,
learning the action schemas of a domain given the predicates, sampled state trajectories, and the structure of the states.
These works however do not address the learning of the predicates.

\section{Formulation of the Learning Problem}

We follow the mathematical formulation proposed by \citeauthor{bonet:ecai2020}, and introduce
then extensions and variations: %. The idea of this formulation is the following:

\begin{enumerate}[$\bullet$]
  \item Given a labeled graph $G \,{=}\,\tup{V,E,L}$, where the nodes $n$ correspond to the different (black box) states,
    and the edges $(n,n')$ in $E$ with label $l \in L$, correspond to state transitions produced by an action with label $l$,
  \item find a ``simplest'' planning instance $P = \tup{D,I}$ such that the state graph $G(P)$ defined by $P$ and $G$
    are isomorphic. %match in a suitable way (are isomorphic).
\end{enumerate}

\begin{figure*}[t]
  \centering
  \begin{tabular}{ccccccc}
    \resizebox{.40\columnwidth}{!}{\includegraphics{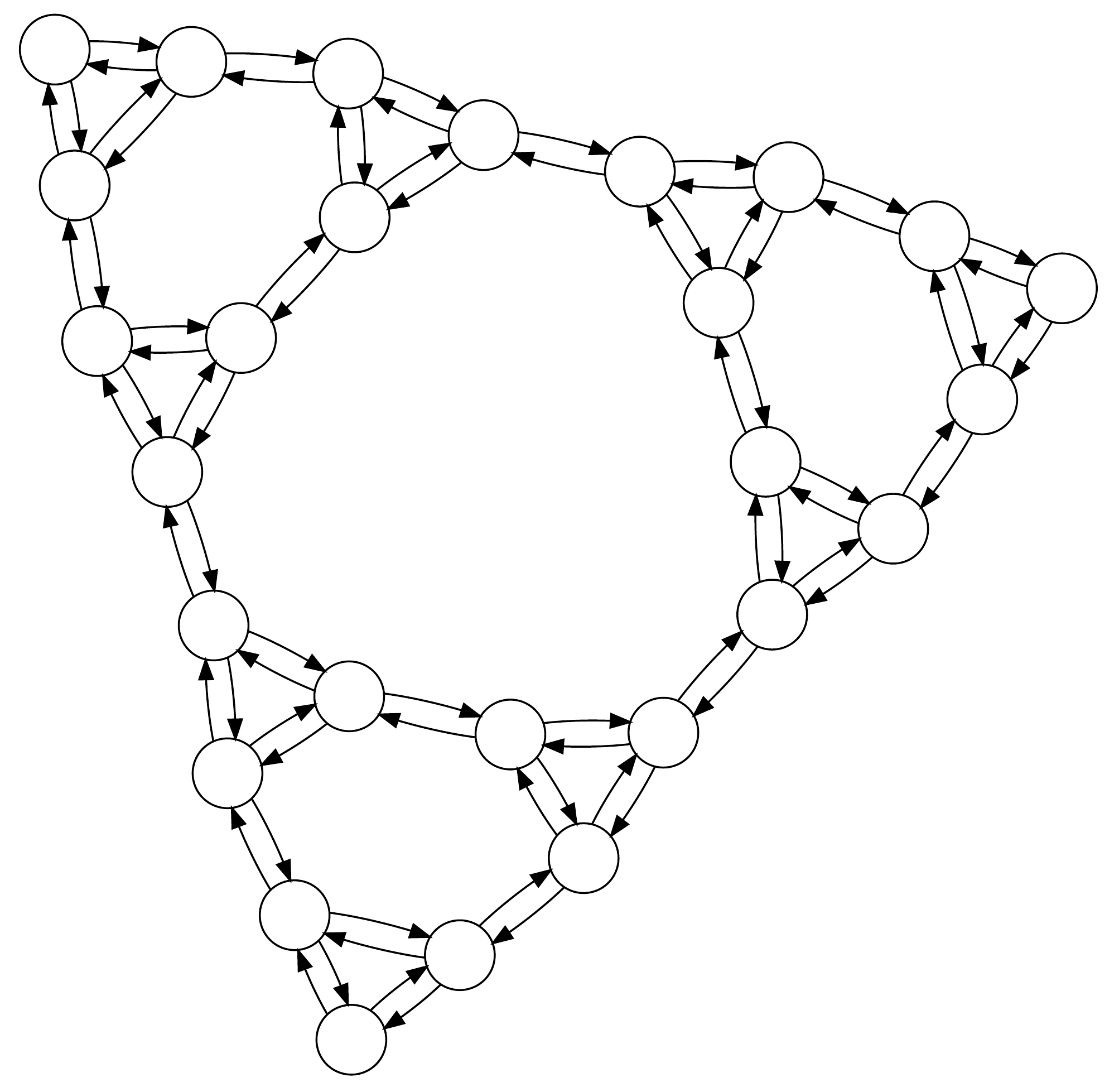}} &&
    \resizebox{.40\columnwidth}{!}{\includegraphics{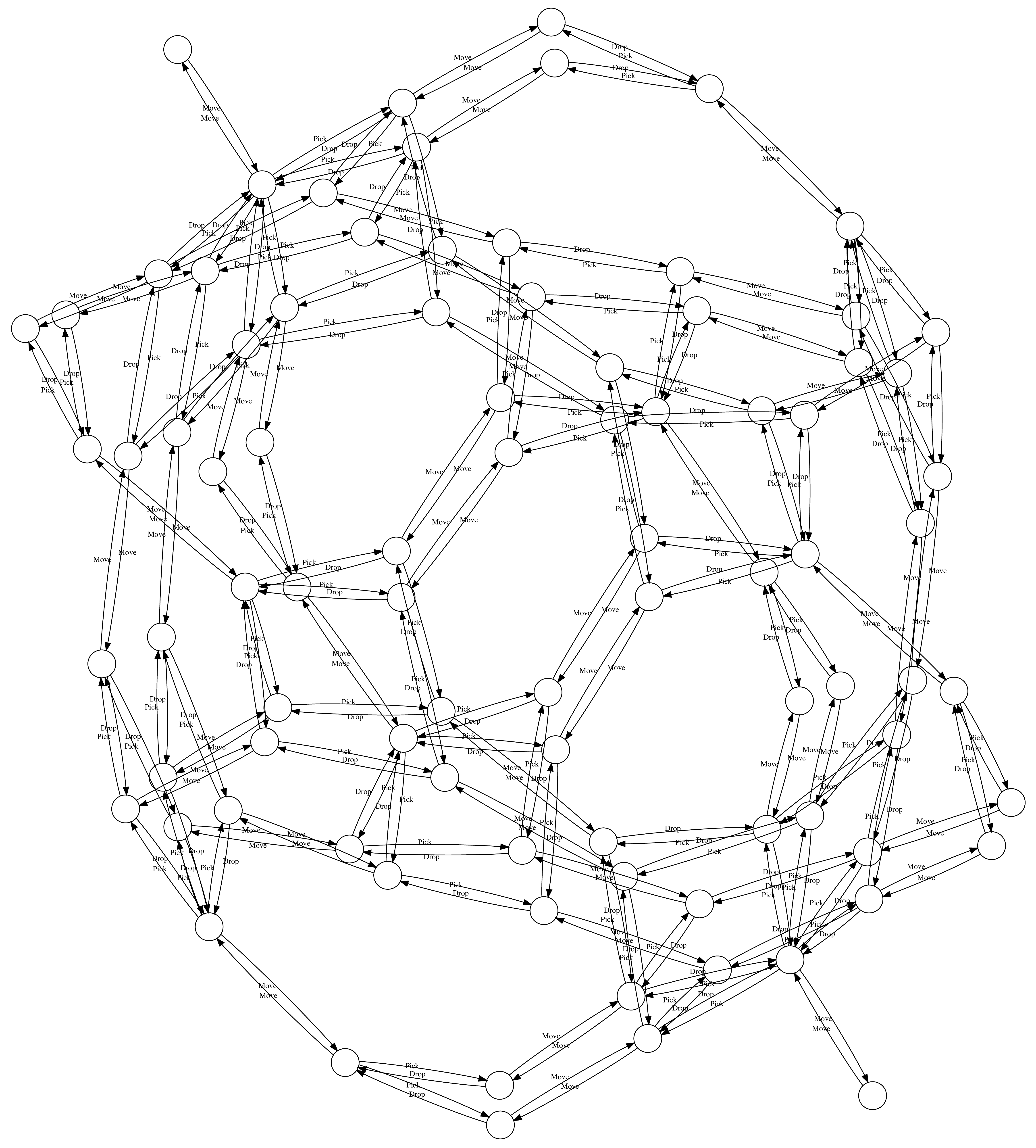}} &&
    \resizebox{!}{.43\columnwidth}{\includegraphics{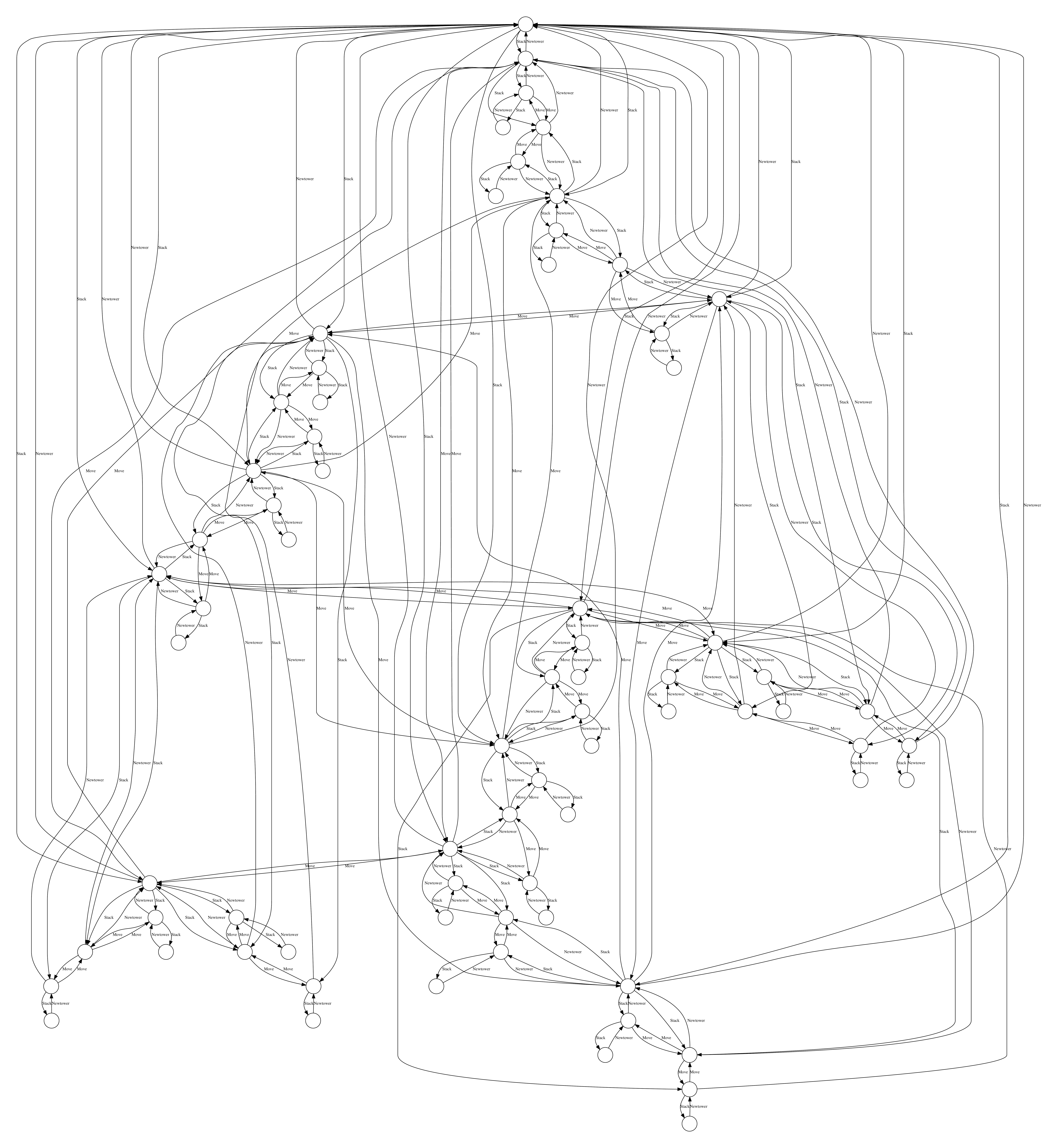}} &&
    \resizebox{!}{.43\columnwidth}{\includegraphics{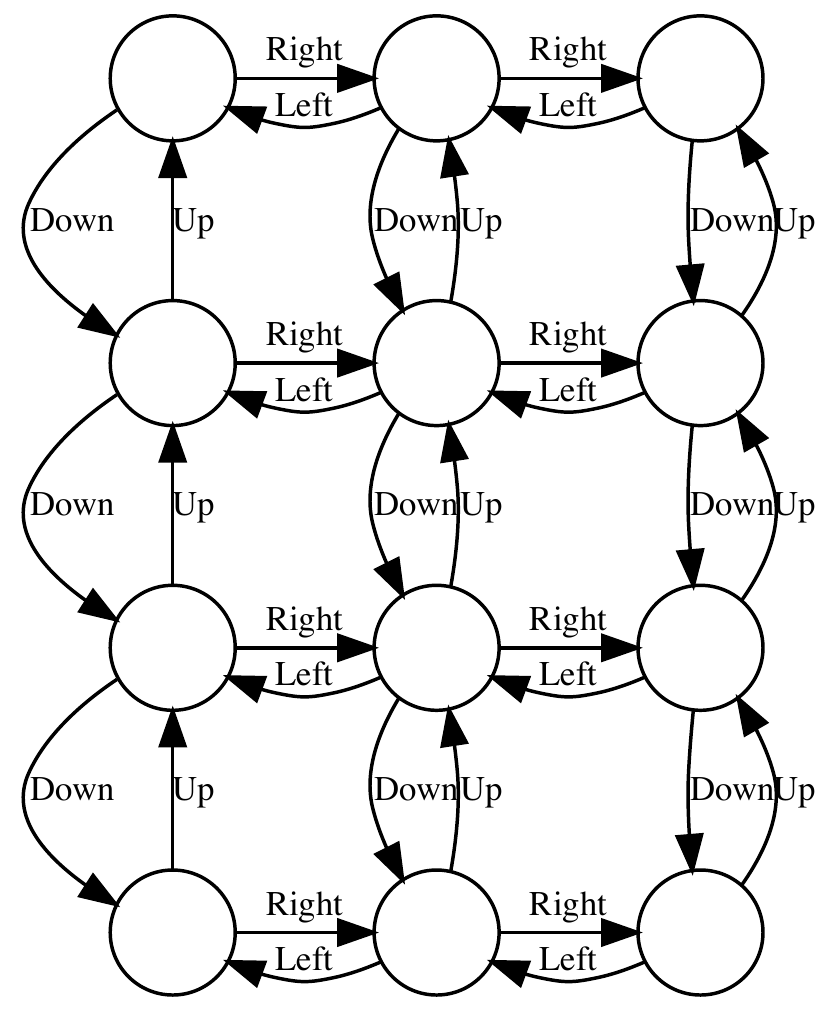}} \\
    {\footnotesize (a) Towers of Hanoi (1 label)} &&
    {\footnotesize (b) Gripper (3 labels)} &&
    {\footnotesize (c) Blocksworld (3 labels)} &&
    {\footnotesize (d) $4{\times}3$ Grid (4 labels)}
  \end{tabular}
  \vskip -.5em
  \caption{Input data for learning representations in 4 planning domains
    from Bonet and Geffner. Graphs can be zoomed in to reveal labels.
    %\textcolor{red}{*** Arrow isn't here ***}
  }
  \label{fig:graphs}
\end{figure*}

Multiple input graphs $G_1, \ldots, G_k$ are handled in a similar way: find simplest common domain $D$ and
instances $P_i\,{=}\,\tup{D,I_i}$ whose state graphs $G(P_i)$ match the given graphs $G_i$.
The formulation assumes that the input graphs $G_i$ are \emph{complete}
in the sense that they do not miss any edge, and that they are \emph{noise-free}
in the sense that no edges in the input graphs are wrong, and that different nodes
stand for different states (different sets of true literals).
%We will relax both assumptions later on.
% 
The input graphs $G_i$ are actually \emph{labeled graphs} with \emph{action labels}
used to add information about the actions in the input.
% In domains where the number of actions
% is fixed and does not change with the instances, the labels can be used to represent the actions,
% yet in most relational domains like Blocksworld, this is not possible as the actions depend on the instance.
% In those cases, however, the action labels in the input can be associated with the names of action schemas.
% E.g., in Blocksworld, the action labels can be four: Stack, Unstack, Pick from table, and Put on table.
These action labels do not convey information about the structure of the action schemas,
nor about their arities or the predicates involved in their preconditions and effects.
Figure~\ref{fig:graphs}, from \citeauthor{bonet:ecai2020}, shows the input graphs considered in their paper
corresponding to instances of the Towers of Hanoi, Gripper, Blocksworld, and Grid.
The graphs displayed are the sole inputs to their representation learning scheme and ours.

\subsection{Formalization: Learning and Verification}

A (classical) planning instance is a pair $P\,{=}\,\tup{D,I}$ where
$D$ is a first-order planning domain and $I$ represents the {instance information}.
The planning domain $D$ contains a set of predicate symbols and a set of action schemas with preconditions
and effects given by atoms $p(x_1, \ldots, x_k)$ or their negations, where $p$ is a domain predicate
and each $x_i$ is a variable representing one of the arguments of the action schema.
The instance information is a tuple $I\,{=}\,\tup{O,Init,Goal}$ where $O$ is a (finite) set of object
names $c_i$, and $Init$ and $Goal$ are sets of ground atoms $p(c_1, \ldots, c_k)$ or their negations.
The actual name of the constants in $O$ is irrelevant and can be replaced by numbers in the interval
$[1,N]$ where $N = |O|$. %is the number of objects in $O$.
Similarly, goals are included in $I$ to keep the notation
consistent with planning practice, but they play no role in the formulation.

A planning problem $P\,{=}\,\tup{D,I}$ defines a \emph{labeled} graph $G(P)\,{=}\,\tup{V,E,L}$
where the nodes $n$ in $V$ correspond to the states $s(n)$ over $P$, and there is an
edge $(n,n')$ in $E$ with label $a$, $(n,a,n')$, if the state transitions $(s(n),s(n'))$
is enabled by a ground instance of the schema $a$ in $P$. The states $s(n)$ are maximally
consistent sets of ground literals over $P$ that comply with: $Init$ corresponds to
state $s(n_0)$ for some node $n_0$, %such that $s(n_0)$ corresponds to $Init$,
states $s(n)$ and $s(n')$ are different if $n \not= n'$, and
$(s(n),s(n'))$ is enabled by a ground action iff the preconditions
of the action are true in $s(n)$, the effects are true in $s(n')$, and
literals whose complements are not made true by the action have
the same value in $s(n)$ and $s(n')$.
%The {learning problem} addressed is then:

\begin{definition}[\citeauthor{bonet:ecai2020}, \citeyear{bonet:ecai2020}]
  The \textbf{learning problem} is to find the simplest instances $P_i\,{=}\,\tup{D,I_i}$
  that account for a set of \textbf{input labeled graphs} $G_i$, $i=1, \ldots, k$.
  \label{def:rdp}
\end{definition}

\vskip -1em

%\textcolor{red}{\bf ** CHECK: 1-1 and onto reqs.\ for $h$ and $g$ **} \\
Here an instance $P$ \textbf{accounts for} a labeled graph $G$ when there is a
1-1 and onto function $h$ between the reachable nodes in $G(P)$ and those in $G$,
and a 1-1 and onto function $g$ between the action labels in $G(P)$ and those in $G$,
such that $(n,a,n')$ is a labeled edge in $G(P)$ iff $(h(n),g(a),h(n'))$ is a labeled edge
in $G$. In words, $P$ accounts for a graph $G$ when the graph $G(P)$ and $G$ match
in this way.

This is a learning problem and not a synthesis problem; namely, the representations $P_i=\tup{D,I_i}$
are not deducible from the input graphs; they are inferred under suitable regularity (simplicity) assumptions,
and the learned domains $D$ are expected to \textbf{generalize} to other larger instances.

Given a suitable definition of ``simplest'', the learning problem becomes a combinatorial optimization
problem, as the space of possible domain representations $D$ is made finite once a bound on
the number of action schemas, predicates, arguments (arities), and objects is defined.
% In every domain, these parameters can be safely assumed to take
% small values, except for the number of objects. In this approach, it is assumed that the number of objects
% in the input graphs (training data) is small as well, as the domain representations are to be learned
% from small examples.
Testing the \textbf{generalization} of the learned domain in other instances
is another combinatorial problem:

\begin{definition}
  The \textbf{verification problem} is to test whether there are instances
  $P'_i\,{=}\,\tup{D,I'_i}$ over learned domain $D$ that account
  for a set of \textbf{testing labeled graphs} $G'_i$, $i=1, \ldots, k'$.
\end{definition}

The verification problem is actually a subproblem of the learning problem where
the domain $D$ is not learned but fixed, and just the instance information needs
to be inferred.

\section{Basic ASP Encoding}

\citeauthor{bonet:ecai2020} \shortcite{bonet:ecai2020} address the learning and verification problems above
as \textbf{SAT problems} using a vector of hyperparameters $\alpha$ to represents the \emph{exact}
number of action schemas and the arity of each one of them, the number of predicate symbols
and the arity of each one of them, and so on. 
% the number of different atoms in the schemas,
% the total number of unary and binary static predicates, and the number of objects.
Since the true value of these parameters is not known, they
% use suitable bounds on them, and
generate a propositional theory $T_\alpha(G)$ for each possible vector of hyperparameters $\alpha$
within certain  bounds, where $G$ is the input graph (a single input graph is shown
to be enough for learning the representations). This number of propositional theories
can be very high. 
% Depending on the number of action labels
% used in $G$, the number of resulting theories range from 6,390 (Hanoi, 1 label),
% to 19,050 (Blocksworld and Gripper, 3 labels) and 37,800 (Grid, 4 labels).
% They
% then sample 10\% of these theories, and pass them to a SAT solver. Even with exact
% hyperparameters, the \emph{average} time for solving each of these theories remains high,
% from 1,400 to 1,800 seconds for Blocks, Hanoi, and Gripper. This is in part because
% the SAT theories are large, somewhat complex, and often UNSAT.

Our move from SAT to ASP is  aimed at  getting  more compact and transparent encodings
that are easier to understand and explore, and that for this reason, can  also
deliver superior performance.
% Fortunately, this has been achieved, after careful testing and evaluation.
We present the resulting encodings in two parts, first
a basic ASP encoding that is easier to follow and understand, and then the
required optimizations. In these encodings, the learning problems considered by \citeauthor{bonet:ecai2020}
are solved by a handful of calls to the ASP solver \clingo \cite{torsten:asp,torsten:clingo},
as opposed to the thousands of calls required in the  SAT encodings. Moreover, once
a preference ordering on solutions is defined, some of these problems are shown to be solved \emph{optimally}, meaning that there is
no simpler representation compatible with the input data.

\subsection{Program}

The ASP code \aspg for learning a first-order instance $P\,{=}\,\tup{D,I}$ from a single input graph $G$
is shown in Figure~\ref{fig:program}. The code can be easily generalized to learn the instances
$P_i\,{=}\,\tup{D,I_i}$ from multiple input graphs $G_i$. The graph $G$ is assumed to be encoded using the
atoms \atom|node(S)| and \atom|tlabel(T,L)| where \atom|S| and \atom|T=(S1,S2)| denote nodes and transitions
in the graph $G$, and \atom|L| denotes the corresponding action label.
The resulting lifted action schemas are encoded via the \atom|prec/3| and \atom|eff/3| atoms (the integer at
the end denotes the arity of the predicate), while the factored states
of the nodes \atom|S|, via the \atom|val/2| and \atom|val/3| atoms, the first for static atoms and
the second for dynamic ones. The {main constraints} are at the bottom of the program in Fig.~\ref{fig:program}: transitions \atom|(S1,S2)| are assigned
exactly to one ground action through the atoms \atom|next(A,OO,S1,S2)| where \atom|OO| are the objects that instantiate
the arguments of the action schema \atom|A|. As a result, the value of the ground atoms in the states \atom|S1| and \atom|S2|
has to be compatible with this ground action. In addition, the ground action \atom|A(OO)| has to be applicable in \atom|S1|,
and if it is applicable in a node \atom|S1'|, then there has to be a node \atom|S2'| such that \atom|next(A,OO,S1',S2')| is true.
Also, the sets of literals true in a node (the states associated to nodes) have to be different for different nodes.
The static predicates and atoms refer to those that control the grounding; i.e., they appear in the precondition of
action schemas but do not appear in the effects, so their value do not depend on the state.
%(and hence \atom|val| atoms for static ground atoms do not include an state argument).

The first part of the program (lines 1--\ref{asp:bounds:end}) sets up the bounds on the domain parameters: max number of predicates (5) and static predicates (2),
and max numbers of effects and preconditions per action schema (6). In addition, the arity of actions is bounded by 3, and the
arity of predicates by 2. The number of action schemas is set to the number of action labels, and the number of objects is
fixed to the constant \atom|num\_objects| that is passed to the solver. In our experiments, the solver is instantiated
with values $1,\ldots,10$ for this parameter; i.e., for each input graph, the solver is called up to 10 times. 
% a number that is orders of magnitude smaller than the number of calls in the SAT approach.
%The reason for treating the number of objects in a different way than the other parameters
%is that when the value for the number of objects is left open, the search becomes less effective.

The second part (lines \ref{asp:lifted:begin}--\ref{asp:lifted:end}) sets up
the action schemas and their lifted preconditions and effects, while the
third part (lines \ref{asp:grounding:begin}--\ref{asp:grounding:end}) encodes
the grounding: for a given number of objects, it generates the possible object tuples \atom|OO|
that can instantiate the arguments of action schemas and predicates. % inside such action schemas.
Pairs \atom|(A,OO)| and \atom|(P,OO)| denote ground actions and
ground atoms. Atoms \atom|map(VV,OO1,OO2)| tell that the variable list
\atom|VV| instantiates to \atom|OO2| for the ground instantiation \atom|OO1|
of the action arguments.
%, with the clauses for \atom|map(Pargs,Args,Objs)| encoding that the lifted
% predicate arguments \atom|Pargs| of an an action schema whose arguments have
% been instantiated to \atom|Args|, should be instantiated to \atom|Objs|.
Truth values \atom|V| of ground atoms \atom|(P,OO)| in the state
\atom|S| are encoded with atoms \atom|val((P,OO),S,V)|, with the state
\atom|S| omitted for static predicates.

The last part (lines \ref{asp:last:begin}--\ref{asp:last:end}) encodes the main constraints:
1)~if the ground action \atom|(A,OO)| is assigned to edge \atom|(S1,S2)|, it must be applicable in \atom|S1|,
the values of the ground literals in \atom|S1| and \atom|S2| must be consistent with the action effects, % and non-effects,
and the action labels must coincide,
2)~if the ground action \atom|(A,OO)| is applicable in a node, it must be associated with a unique outgoing edge, and
3)~the sets of true ground literals for different nodes must be different. This fragment of the program \aspg  makes
an \emph{assumption} which is most often but not always true; namely, that if  two ground instances of an action schema are
applicable in a  state, their  effects will not be the same.\footnote{An schema with arguments $x$ and $y$, 
effects $p(x)$ and $p(y)$, and no preconditions, for example,  would violate  this assumption. Thanks to Andres Occhipinti for point this out.}
Provided with  this assumption, the correctness of the encoding can be expressed as:
%(** need rephrasing and extra conditions, and proof, although not in submission. Define $ASP(G)$. ***)
%% Hector changed above 21/5

\begin{theorem}
  Let $G$ be an input graph encoded with the \atom|node/2| and \atom|tlabel/2| atoms.
  Then $M$ is an answer set of the program \aspg iff
  there is a planning instance $P\,{=}\,\pair{D,I}$ that accounts for $G$
  and is compatible with the given bounds; $D$ and $I$ can be read off from $M$.
\end{theorem}

\begin{figure*}
  \begin{Verbatim}[gobble=4,frame=single,numbers=left,codes={\catcode`$=3}]
    \textcolor{blue}{% Constants}
    #const max_predicates=5.
    #const max_static_predicates=2.
    #const max_effects=6.           % for each action
    #const max_precs=6.             % for each action

    \textcolor{blue}{% Actions, predicates, static predicates and objects}
    action(L) :- tlabel(T,L).  \{ a_arity(A,1..3) \} = 1 :- action(A).
     pred(1..max_predicates).  \{ p_arity(P,1..2) \} = 1 :-   pred(P).
    p_static(max_predicates-max_static_predicates+1..max_predicates).
    object(1..num_objects).\label{asp:bounds:end}

    \textcolor{blue}{% Tuples of variables for lifted effects and preconditions}\label{asp:lifted:begin}
    argtuple((V1,  ),1) :- V1=1..3.
    argtuple((V1,V2),2) :- V1=1..3, V2=1..3.

    \textcolor{blue}{% Generate lifted preconditions and effects (at least 1) of action schemas}
!% at least one effect? (yes if there are no 0-length loops) excludes static predicates
      \{ prec(A,(P,T),0..1) : p_arity(P,AR), argtuple(T,AR)                  \}   max_precs :- action(A).
    1 \{  eff(A,(P,T),0..1) : p_arity(P,AR), argtuple(T,AR), not p_static(P) \} max_effects :- action(A).

    \textcolor{blue}{% Check that variables mentioned in precs and effects are action arguments}
    :-  eff(A,(_,(V, )),_), a_arity(A,ARITY), ARITY < V.
    :-  eff(A,(_,(V,_)),_), a_arity(A,ARITY), ARITY < V.
    :-  eff(A,(_,(_,V)),_), a_arity(A,ARITY), ARITY < V.
    :- prec(A,(_,(V, )),_), a_arity(A,ARITY), ARITY < V.
    :- prec(A,(_,(V,_)),_), a_arity(A,ARITY), ARITY < V.
    :- prec(A,(_,(_,V)),_), a_arity(A,ARITY), ARITY < V.

    \textcolor{blue}{% Tuples of objects for grounding the action schemas and atoms}
    objtuple((O1,     ),1) :- object(O1).
    objtuple((O1,O2   ),2) :- object(O1), object(O2).
    objtuple((O1,O2,O3),3) :- object(O1), object(O2), object(O3).\label{asp:lifted:end}

    \textcolor{blue}{% Possible values of ground atoms in the states associated to nodes}\label{asp:grounding:begin}
    \{ val((P,OO),  0..1) \} = 1 :- p_arity(P,ARITY),     p_static(P), objtuple(OO,ARITY).
    \{ val((P,OO),S,0..1) \} = 1 :- p_arity(P,ARITY), not p_static(P), objtuple(OO,ARITY), node(S).

    \textcolor{blue}{% Map selects grounding of lifted atoms in schema from grounding of action arguments}
    map( (1,),     (O1,),  (O1,)) :-      objtuple((O1,),1).
    map( (1,),   (O1,O2),  (O1,)) :-    objtuple((O1,O2),2).
    ...
    map((1,1),     (O1,),(O1,O1)) :-      objtuple((O1,),1).
    map((1,1),   (O1,O2),(O1,O1)) :-    objtuple((O1,O2),2).
!map((1,2),   (O1,O2),(O1,O2)) :-    objtuple((O1,O2),2).
!map((2,1),   (O1,O2),(O2,O1)) :-    objtuple((O1,O2),2).
!map((2,2),   (O1,O2),(O2,O2)) :-    objtuple((O1,O2),2).
!map((1,1),(O1,O2,O3),(O1,O1)) :- objtuple((O1,O2,O3),3).
!map((1,2),(O1,O2,O3),(O1,O2)) :- objtuple((O1,O2,O3),3).
!map((1,3),(O1,O2,O3),(O1,O3)) :- objtuple((O1,O2,O3),3).
!map((2,1),(O1,O2,O3),(O2,O1)) :- objtuple((O1,O2,O3),3).
!map((2,2),(O1,O2,O3),(O2,O2)) :- objtuple((O1,O2,O3),3).
!map((2,3),(O1,O2,O3),(O2,O3)) :- objtuple((O1,O2,O3),3).
!map((3,1),(O1,O2,O3),(O3,O1)) :- objtuple((O1,O2,O3),3).
!map((3,2),(O1,O2,O3),(O3,O2)) :- objtuple((O1,O2,O3),3).
!map((3,3),(O1,O2,O3),(O3,O3)) :- objtuple((O1,O2,O3),3).
!map((1,1),(O1,O2,O3),(O1,O1)) :- objtuple((O1,O2,O3),3).
!map((1,2),(O1,O2,O3),(O1,O2)) :- objtuple((O1,O2,O3),3).
!map((1,3),(O1,O2,O3),(O1,O3)) :- objtuple((O1,O2,O3),3).
!map((2,1),(O1,O2,O3),(O2,O1)) :- objtuple((O1,O2,O3),3).
!map((2,2),(O1,O2,O3),(O2,O2)) :- objtuple((O1,O2,O3),3).
!map((2,3),(O1,O2,O3),(O2,O3)) :- objtuple((O1,O2,O3),3).
    ...
!map((3,1),(O1,O2,O3),(O3,O1)) :- objtuple((O1,O2,O3),3).
    map((3,2),(O1,O2,O3),(O3,O2)) :- objtuple((O1,O2,O3),3).
    map((3,3),(O1,O2,O3),(O3,O3)) :- objtuple((O1,O2,O3),3).

    \textcolor{blue}{% Check preconditions: ground action A(OO1) is applicable in node S}
    appl(A,OO1,S) :- action(A), a_arity(A,ARITY), objtuple(OO1,ARITY), node(S),
                     val((P,OO2),  V) : prec(A,(P,T),V), map(T,OO1,OO2),     p_static(P);
                     val((P,OO2),S,V) : prec(A,(P,T),V), map(T,OO1,OO2), not p_static(P).

    \textcolor{blue}{% If ground action A(OO) applicable in S1, assigned to some edge (S1,S2) with same label}\label{asp:partial1:begin}
    \{ next(A,OO,S1,S2) : tlabel((S1,S2),A) \} = 1 :- appl(A,OO,S1).\label{asp:grounding:end}

    \textcolor{blue}{% Every edge is assigned to a ground action with the same label}\label{asp:last:begin}
    :- tlabel((S1,S2),A), \{ next(A,OO,S1,S2) \} $\neq$ 1.\label{asp:partial1:end}

    \textcolor{blue}{% Effects and inertia}\label{asp:partial2:begin}
!% if (S1,A,S2) with O1 in V1 and O2 in V2,% then the effects must hold in S2
    :- eff(A,(P,T),V), next(A,OO1,S1,S2), map(T,OO1,OO2), val((P,OO2),S2,1-V).
    :- tlabel((S1,S2),_), val(K,S1,V), val(K,S2,1-V), not caused(S1,S2,K).
    caused(S1,S2,(P,OO2)) :- eff(A,(P,T),V), next(A,OO1,S1,S2), map(T,OO1,OO2).\label{asp:partial2:end}
!%
!% Another option for inertia with a single rule:
!% if A(OO1) is applicable in S1 and is connected to S2,
!% if P(OO2) changes its value
!% then there must be some effect accounting for the change
!% :- next(A,OO1,S1,S2), val((P,OO2),S1,V), val((P,OO2),S2,1-V),
!%    not eff(A,(P,T),1-V) : map(T,OO1,OO2).

    \textcolor{blue}{% Different nodes are different states}
    :- node(S1), node(S2), S1 < S2, val((P,T),S2,V) : val((P,T),S1,V).\label{asp:last:end}\label{asp:partial:end}
!%
!% Display
!#show eff/3.
!#show prec/3.
!#show labelname/2.
!#show val(K,S) : val(K,S,1).
!#show val(K  ) : val(K,  1).
!#show object/1.
  \end{Verbatim}
  %\vskip -1em
  \caption{Base code of ASP program \aspg for learning a first-order planning instance $P\,{=}\,\tup{D,I}$
    from a single input graph $G$. The graph $G$ is assumed to be encoded with atoms \protect\atom|node(S)| and
    \protect\atom|tlabel(T,L)| %$node(S)$ and $tlabel(T,L)$   atoms
    where \protect\atom|T=(S1,S2)| stands for the transitions in the graph and \protect\atom|L| for the corresponding
    action label. See text for explanations. %** Problem compiling ``atoms'' **
  }
  \label{fig:program}
\end{figure*}

\section{Extensions and Optimization}

The code shown in Fig.~\ref{fig:program} captures a correct encoding of the learning problem
but it is not optimized for performance. We discuss next some variations and extensions. 
The code for  these extensions is shown  in Fig.~\ref{fig:program:ext}.
%starting with the optimization criterion that is used
%for preferring some models (answer sets) to others.

\subsubsection{Optimization.}
The program $\aspg$ for a given graph $G$ usually has many models.
Following Occam's razor, the models that are simpler are more likely to result
in action schemas that generalize to larger instances and hence to pass
the verification test.
Models are thus ranked using four numerical criteria ordered \textbf{lexicographically}
from most important to less important as follows:
A)~sum $N_a$ of the action schemas arities,
B)~sum $N_p$ of the arities of  non-static predicates,
C)~sum $N_s$ of the arities of static predicates,
D)~maximum number $N_g$ of true  ground atoms per state.

\Omit{
\begin{enumerate}[1.]
  \item Models with action schemas of smaller arities are preferred, by
    computing the sum $N_a$ of the action schemas arities.
  \item Models with dynamic predicates of smaller arities are preferred,
    by computing the sum $N_p$ of the arities of  non-static predicates.
  \item Models with static predicates of smaller arities are preferred,
    by computing the sum $N_s$ of the arities of static predicates.
  \item Models that minimize the number of true atoms at states (i.e.,
    the size of state) is preferred, by computing the max number $N_g$ of true
    ground atoms per state.
\end{enumerate}
}

%Lines \ref{asp:opt:begin}--\ref{asp:opt:end} in Figure~\ref{fig:program:ext} implement the optimization criterion.
If $V(M)=(N_a,N_p,N_s,N_g)$ is the resulting numerical cost vector for model $M$,
and $V(M')=(N'_a,N'_p,N'_s,N'_g)$ is the vector for model $M'$, $M$ is preferred to $M'$
if $V(M)$ is lexicographically  smaller than $V(M')$.
% in one of the four dimensions, and it is no greater
% in any of the preceding ones.
A model $M$ is \textbf{optimal} or \textbf{simplest} if there
is no other $M'$ preferred to $M$. This exact optimization criterion is given to \clingo
along with the code (lines \ref{asp:opt:begin}--\ref{asp:opt:end} in Fig.~\ref{fig:program:ext}).
%
%  \clingo combines different techniques for finding optimal models  (branch and bound and unsatisfiable core),
% and enumerates models such that every one is better than the previous ones.
%
% Javier: This is not the case by default, but can be activated using option --opt-strategy=bb,hier
%\clingo then looks for models that yield optimal values for the dimension $N_a^*$,
%then for models that yield optimal values for the second dimensions $N_p^*$ given $N_a^*$,
%and so on.
%
For a given \textbf{time window}, \clingo may terminate earlier with an optimal solution,
or run out of time, returning the \textbf{best solution} found that far.
% We usually report
% the first solution, the best solution, and whether this solution was proved to be optimal.

\Omit{
We make no claim that the optimal criterion above is the most suitable; other similar criteria
can be used. We will see that many of the problems considered by \citeauthor{bonet:ecai2020}
can be solved optimally in this way, producing representations that generalize well.
This, however, is not a requirement, and indeed, in many cases, we will see that
good but non-optimal solutions also generalize well. On the other hand, representations that
comply with the bounds but are much larger than needed often do not, although this could
be avoided through the use of more training data.
}

\Omit{
\subsubsection{Number of true atoms per state.}
The least significant optimization criterion involves the number $N_g$ that bounds the
number of ground atoms made true at each state. The program enforces such a bound $N_g$
in lines \ref{asp:states:begin}--\ref{asp:states:end} in Fig.~\ref{fig:program:ext}.
}

\subsubsection{Invariants.}
One way to speed up SAT and ASP solvers is by adding implicit constraints. 
State invariants  express formulas that are true in all (reachable) states,
and they are implied by the truth valuation of the initial state and the structure
of the action schemas. One particular type of invariants is given by sets of atoms $R$
such that \emph{exactly-1} of the atoms in $R$ is true in every state. An invariant
of this form, for example in  Blocksworld without an arm, is given
by the sets of atoms $\{on(x,y):y\}\,\cup\,\{ontable(x)\}$ for any block $x$,
where $y$ ranges over all blocks.
The variable $x$ is  the \textbf{free variable} of the \textbf{invariant schema},
while other variables like $y$ range over their possible instantiations.

We extend the basic ASP encoding so that \emph{exactly-1} invariant schemas can be
constructed and enforced automatically during the search. This is achieved by introducing atoms
\atom|schema(N,P,I)| where \atom|N| is an index over a max number of lifted invariants,
\atom|P| is a predicate, and \atom|I| is the index of the argument of $P$ that represents
the free variable if $P$ is binary (for unary predicates there is no choice).
The solver is free to choose the atoms \atom|schema(N,P,I)| that are true,
and the resulting constraints are enforced as invariants; i.e., they must be true
in all states. Furthermore, to force some invariants to be true, every \emph{dynamic binary predicate}
is constrained to be part of some invariant. This is a common situation
where dynamic binary predicates like $p(x,y)$ are used to encode multivalued
state variables. %: ``the $p_2$ of $x$'' and their possible $y$ values, or ``the $p_1$ of $y$ and their possible $x$ values.
%%These invariants, however, would not be needed if rather than looking for representations in languages
%%such as STRIPS (with negation), we would be looking for representations in more expressive languages
%%such as Functional STRIPS \cite{geffner:fstrips}.
Lines \ref{asp:inv:begin}--\ref{asp:inv:end} in Fig.~\ref{fig:program:ext} implements
the construction and enforcement of invariants.
%%%\textcolor{red}{**** CHECK ****}
%%%The use of static atoms in invariants (Fig.~\ref{fig:invariants})
%%%allow for invariants like $\{at(x,r), hold(x)\}$ for the free variable $x$, where $r$
%%%ranges over the rooms in Gripper, when $x$ is of type Ball. The resulting \emph{exactly-1} invariant
%%%is $\{at(x,r),hold(x),\neg ball(x)\}$.

%% Removed for lack of space; doesn't seem critical and one page with code seems enough

\subsubsection{Other extensions.}
The actual code of the program \aspg incorporates other extensions
like constraints for symmetry breaking, special rules for
handling cycles of size two in the input graph, and transformations for treating all action
schemas (resp.\ predicates) as if they are all of the same (max) arity (to reduce
grounding size). %  and a number of heuristics for the solver expressed in the language of \clingo.
% For lack of space, we cannot describe these other extensions,
The code and the relevant data will be made public.

\begin{figure*}[t]
  \begin{Verbatim}[gobble=4,frame=single,numbers=left,codes={\catcode`$=3}]
    \textcolor{blue}{% Optimization}\label{asp:opt:begin}
    #minimize \{ 1+N@4, A : a_arity(A, N)                  \}.
    #minimize \{ 1+N@3, P : p_arity(P, N), not p_static(P) \}.
    #minimize \{   N@2, P : p_arity(P, N),     p_static(P) \}.
    #minimize \{   N@1, N : state_bound(N)                 \}.\label{asp:opt:end}
!
!   \textcolor{blue}{% Heuristics}\label{asp:heuristic:begin}
!   #heuristic   a_arity(A,ARITY). [1,level]
!   #heuristic   p_arity(A,ARITY). [1,level]
!   #heuristic         eff(A,M,V). [1,level]
!   #heuristic        prec(A,M,V). [2,level]
!   #heuristic         val(K,S,V). [1,level]
!   #heuristic   next(A,OO,S1,S2). [1,level]\label{asp:heuristic:end}

    \textcolor{blue}{% Bound number of true atoms per state}\label{asp:states:begin}
    #const max_true_atoms_per_state=10.
    \{ state_bound(1..max_true_atoms_per_state) \} = 1.
    \{ val((P,OO),S,1) : pred(P), not p_static(P), objtuple(OO,2) \} N :- node(S), state_bound(N).\label{asp:states:end}
!
!   \textcolor{blue}{% Constants}
!   #const max_num_invariants=0.
!   #const must_have_binary_predicates=0.

    \textcolor{blue}{% Choose number and type of invariants}\label{asp:inv:begin}
    #const num_invariants=1.
!   #heuristic invariant(N) : N=1..max_num_invariants. [N, true]
!   invariant(1) :- max_num_invariants $\neq$ 0.
!   \{ invariant(N) \} :- invariant(N-1), N <= max_num_invariants.
    inv(1..num_invariants).

    \textcolor{blue}{% Schemas for invariants}
    \{ schema(N,P,1) \} :- inv(N), pred(P), p_arity(P,1).
    1 \{ schema(N,P,2..3) : pred(P), p_arity(P,2), not p_static(P) \} :- inv(N), bin_preds.
    inv_non_empty(N) :- inv(N), pred(P), schema(N,P,1..3).
    bin_preds :- pred(P), p_arity(P,2), not p_static(P).
!   :- not some_binary_predicate, must_have_binary_predicates $\neq$ 0.
!
!   \textcolor{blue}{% Different invariants}
!   :- schema(N1,P,X) : schema(N2,P,X); invariant(N1), invariant(N2), N1 $\neq$ N2.

    \textcolor{blue}{% Each non-static binary predicate must appear in some invariant}
    inv_used_pred(P) :- pred(P), p_arity(P,2), not p_static(P), inv(N), schema(N,P,2..3).
    :- pred(P), p_arity(P,2), not p_static(P), not inv_used_pred(P).

    \textcolor{blue}{% Enforce invariants at states}
    \{ ! val((P,(O2, O2)), S, 1) : object(O2), schema(N, P, 0), not p_static(P);   % unary { P(O2) : O2 }
 val((P, (O,O)),  1) :             schema(N,P,1),     p_static(P); \textcolor{blue}{% \{ P(O) \} (static)}
       val((P, (O,O)),S,1) :             schema(N,P,1), not p_static(P); \textcolor{blue}{% \{ P(O) \}}
       val((P,(O,O2)),S,1) : object(O2), schema(N,P,2), not p_static(P); \textcolor{blue}{% \{ P(O,O2) : O2 \}}
       val((P,(O2,O)),S,1) : object(O2), schema(N,P,3), not p_static(P)  \textcolor{blue}{% \{ P(O2,O) : O2 \}}
    \} = 1 :- inv(N), inv_non_empty(N), object(O), node(S).\label{asp:inv:end}
!
!   \textcolor{blue}{% Schema for exactly-1 invariants}
!   \{ inv_schema(N, P, 1) \} :- invariant(N), pred(P), p_arity(P, 1).
!   1 \{ inv_schema(N, P, 2..3) : pred(P), p_arity(P, 2), not p_static(P) \} :- invariant(N),  \textcolor{red}{???}
!
!   \textcolor{blue}{% Enforce exactly-1 invariants}
!\{%% val((P,(O2, O2)), S, 1) : object(O2), inv_schema(N, P, 0), not p_static(P);   % unary { P(O2) : O2 }
!\{ val((P,  (O, O)),    1) :             inv_schema(N, P, 1),     p_static(P);
!   \{ val((P,  (O, O)),    0) :             inv_schema(N,P,1),     p_static(P);
!% (static) unary { P(O) }
!     val((P,  (O, O)), S, 1) :             inv_schema(N,P,1), not p_static(P);
!% unary { P(O) }
!     val((P, (O, O2)), S, 1) : object(O2), inv_schema(N,P,2), not p_static(P);
!% binary { P(O,O2) : O2 }
!     val((P, (O2, O)), S, 1) : object(O2), inv_schema(N,P,3), not p_static(P)
!% binary { P(O2,O) : O2 }
!   \} = 1 :- invariant(N), inv_non_empty(N), object(O), node(S).
!
!
!
!
!   \textcolor{blue}{% Additional constraints}
!
!   \textcolor{blue}{% No more changes than the max of effects}
!   :- edge((S1,S2)), max_effects+1 #count\{K : val(K,S1,V), val(K,S2,1-V) \}.
!
!   \textcolor{blue}{% If A has N outgoing edges in S1, then it should be applicable N times}
!   :- node(S1), action(A), N = \{ tlabel((S1,S2),A) \}, \{ appl(A,OO,S1) \} $\neq$ N.
!
!   \textcolor{blue}{% If there are N edges (in total), there should be N applicable actions}
!   :- N = \{ tlabel(E,L) \}, \{ appl(A,B,C) \} $\neq$ N.
!
!   \textcolor{blue}{% If P doesn't have arity 2, the atoms P(O1,O2) where O1 $\neq$ O2 are false}
!   :- val((P,(O1,O2)),S,1), not p_arity(P,2), O1 != O2.
!
!   \textcolor{blue}{% If P doesn't have arity 1 or 2, the atoms P(O1,O2) where O $\neq$ 0 are false}
!   p_arity_1_or_2(P) :- p_arity(P,ARITY), ARITY = 1..2.
!   :- val((P,(O,O)),S,1), not p_arity_1_or_2(P), O $\neq$ 0.
!
!   \textcolor{blue}{% If P doesn't have arity 0, P(0,0) is false}
!   :- val((P,(0,0)),S,1), not p_arity(P,0).
!
!   \textcolor{blue}{% By the previous constraints, if P has no arity, all P(O1,O2) are false}
!
!   \textcolor{blue}{% Define root(0) by default if no root node specified}
!   root(0) :- not root(N) : node(N), N $\neq$ 0.    \textcolor{red}{*** CHECK THIS ***}
  \end{Verbatim}
  %\vskip -1em
  \caption{Fragments of ASP code for different extensions of the basic ASP code.}
  %. \textcolor{red}{*** Blai: REMOVE THIS CODE; too much space. Not needed. Can't Omit it though! ... ***}}
  \label{fig:program:ext}
\end{figure*}

\section{Experimental Results}

We test the performance of the program \aspg with all the extensions above.
The program accepts a single graph $G$ and outputs a model (solution) %(answer set)
from which a first-order planning instance $P=\tup{D,I}$ that matches $G$
%We performed experiments to test the performance of the
%program \aspg with all the extensions above. The program
%accepts a single graph $G$ in the experiments and outputs
%a model (answer set) from which a first-order
%planning instance $P=\tup{D,I}$ that matches $G$
can be read off.

% We report first-time solutions, corresponding to the
% first models found by \clingo, and best-solutions, corresponding
% to the best models found in the given time window. We indicate
% when the best models are proved to be optimal by the solver
% in the sense defined above.
%In those cases, the solver does not
%consume the entire time window and the actual time is shown.

We consider  five domains:  two  versions of Blocksworld (with and without an arm),
Towers of Hanoi, Gripper and Grid;  all from \citeay{bonet:ecai2020},  except Blocksworld
with an arm.
%We considered five domains, Arrow, Blocksworld, Towers of Hanoi, Grid, and Gripper,
%the last four from Bonet and Geffner \shortcite{bonet:ecai2020}.
%Figure~\ref{fig:graphs} shows the input graphs $G$ used in each case. 
The experiments were performed on Amazon EC2's \atom|c5n.12xlarge| nodes
with a limit of 16Gb of memory, 2 hours   for learning,
and 1 hour for verification. Recall that   verification is  a
combinatorial problem similar to the learning problem but 
with  the learned model known and fixed (Section~3.1). 
The data about the instances used is shown in Table~\ref{table:graphs}.
%%%The number of nodes in the graphs range from $8$ to $88$, and the number of
%%%state transitions, from $24$ to $280$.
%%%The number of objects used for generating these instances is
%%%shown next to the domain names and the number of action labels
%%%is shown in a separate column. For example, the input graph
%%%for Hanoi uses 3 disks and 3 pegs, i.e., 6 objects in total,
%%%and all edges have the same action label (``moves from peg to peg'').
%%%The input graphs for Gripper in turn involve 2 or 3 balls that have to be moved
%%%from one to another with a robot with two (???) grippers,
%%%involves *** objects (balls, rooms, grippers***?), 
%%%and three action labels (moves, drops, and pick ups).
%%%Interestingly, the instances learned form these graphs
%%%often involve less number of objects but still generalize well to
%%%larger instances. The resulting encodings, however, are not so
%%%easy to understand, although they can be fully understood,
%%%as opposed to the models learned in general with deep lerning.
For each instance $P$ in the table, the  graph $G(P)$ is used to learn a first-order model, and the
learned  model (if any) is then verified on the other instances $P'$ of the same domain that are listed in the table.
The solver \clingo was  run with 8 threads. 

\begin{table}[t]
  \centering
  \resizebox{\columnwidth}{!}{
\begin{tabular}{@{}lrrrr@{}}
    \toprule
    %                     & \multicolumn{3}{c}{Input TS} && \multicolumn{5}{c}{Statistics SAT Calls} && \multicolumn{4}{c}{Theory for SAT Tasks (avg.)} \\
    % \cmidrule{2-4}\cmidrule{6-10}\cmidrule{12-15}
    Domain                             & \#labels & \#obj & \#nodes & \#edges \\
    \midrule
    %%Arrow 4                          &        1 &     3 &       8 &      24 \\
    %%Arrow 5                          &        1 &     4 &      16 &      64 \\
    %
    blocks1-2 (arm, 2 blocks)          &        4 &     2 &       5 &       8 \\
    blocks1-3 (arm, 3 blocks)          &        4 &     3 &      22 &      42 \\
    blocks1-4 (arm, 4 blocks)          &        4 &     4 &     125 &     272 \\
    blocks1-5 (arm, 5 blocks)          &        4 &     5 &     866 &   2,090 \\
    \midrule
    blocks2-2 (2 blocks)               &        3 &     2 &       3 &       4 \\
    blocks2-3 (3 blocks)               &        3 &     3 &      13 &      30 \\
    blocks2-4 (4 blocks)               &        3 &     4 &      73 &     240 \\
\bottomrule
  \end{tabular}
  }
  %\vskip -.5em
  \caption{Data for the graphs $G(P)$ of the instances $P$ used for learning the domains and for their  verification: 
 numbers of action labels, nodes, and edges. Number of objects in $P$ shown as well.
  }
  \label{table:graphs}
\end{table}

\begin{table}[t]
  \centering
  \resizebox{\columnwidth}{!}{
    \begin{tabular}{@{}lrrrrrr@{}}
      \toprule
             &       &           & \multicolumn{2}{c}{Time to find model} \\
      \cmidrule{4-5}
      %domain & \#obj & \#threads & \#models & first & best & opt \\
      Domain & \#obj & \#mod & First & Best & Ver & Opt \\
\bottomrule
    \end{tabular}
  }
  %\vskip -.5em
  \caption{Results of  \aspg on  the instances in Table~\ref{table:graphs}.
     Columns show the instance used for learning, the number of objects found, and times for the first and best solutions found, and whether the best
    solution verified and was proved to be optimal (simplest). The column ``\#mod'' shows the number of models found in the way to 
    the best model. Times are in seconds.
    % and learning times out after 2 hours. 
    %%%Verification times out after 3600 seconds  and only runs that result in domains that verify (generalize to larger instances) are shown. 
    %%%Verified  models found for most domains, even for those, like the last three domains, where the models  could not be shown to  be the simplest.
    % Results obtained with \clingo running with 8 threads.
  }
  \label{table:base}
\end{table}

Table~\ref{table:base} shows the results of the ASP-based learner on these instances. 
As mentioned above, the solver is called multiple times with different number of objects, from 1 up to 10,
and the models with the  smallest number of objects that passed the verification are reported for each domain
(except for Grid  where models with a smaller number of objects than expected also generalized).
Bounds on the number of action schemas, predicates and their arities are those
expressed in the code shown in Figure~\ref{fig:program}.
The columns ``Ver'' and ``Opt'' in the table refer to whether the model found
verified (generalized to the other instances in Table~\ref{table:graphs}),
and whether it was proved to be optimal (i.e., simplest according to the optimization criterion). 
% For example, the model for blocks2-4 was verified
% on the instances for 2, 3 and 5 blocks, and it was shown to be optimal (simplest).
The table also reports the time to find a model for each instance (time to first model), and
the time  to find the best model  within the time window (2h). A time of 7,200 seconds indicates
that the best model was not proved to be optimal in the time window. The table also displays 
the number of models found by the solver in the optimization process, each model being
 better than the previous one.

The key  observation that can be drawn from  Table~\ref{table:base} is that models that generalize to  the other instances
of the same domain are found in  all of the domains, except in Blocks-1 (Blocks with an arm).
Moreover, some models that are shown to be optimal do not  generalize (Blocks-1-3, Blocks-2-3), while models that
are not proved to be optimal, do (Gripper, Hanoi). The reason for the former is that the  instances
used for learning are too small, which lead to action schemas that do not apply to larger instances.
The reason for the latter is that strict optimality is not a condition for generalization. 
In the case of the Blocks-1 domain, larger instances like Blocks-1-4 timed out. Regarding the size
of the ground programs, they feature  between 30K to 70K rules and constraints
in  Grid and Gripper-2, and  between 1M to 2M  in Blocks, Gripper-3 and Hanoi.
Similarly, the number of variables ranges between 50K to 165K in the former, 
and between 300K to 800K in the latter. 

The improvements in relation to the results reported by \citeay{bonet:ecai2020}
using a SAT approach are significant. As it was mentioned before,
for each input $G$,
% they consider propositional theories  $T_\alpha(G)$ where $\alpha$ is a vector of hyperparameters
% that comply with given bounds .
% that represents the \emph{exact}
% number of action schemas and the arity of each one of them, the number of predicate symbols
% and the arity of each one of them, and so on.
they need to call the SAT solver over the theory  $T_\alpha(G)$ for each possible vector of hyperparameters $\alpha$, which 
means a number of calls that range from 6390 (Hanoi, 1 label) to 19,050 (Blocksworld and Gripper, 3 labels)
and 37,800 (Grid, 4 labels), which are not done exhaustively. In the ASP approach, 
\aspg is called a maximum number of times that is given by a bound on the number of objects,
and in many cases, the models found are shown  to be optimal.
% This improvement does not mean
% that ASP and \clingo technology is more efficient than SAT technology but that they offer
% a broader range of possibilities for encodings our learning problem that is subtle.
% as it involves finding a first-order symbolic language to describe plain graphs.
% Moreover by being a very high-level modeling language, \clingo allows us to explore and evaluate
% these possibilities at a high level as well.
Among the extensions of the base code that have the greatest impact on performance, there are two: learning invariants while forcing dynamic
predicates to appear in an invariant, and calling the solver separately for each number of
objects as opposed to letting the solver search for this value.
%The impact of the other code extensions code is smaller.

\subsection{Examples of Learned Representations}

As in the work of \citeay{bonet:ecai2020}, the learned representations are often more
succinct than the hand-crafted representations, usually using the same predicate to
represent different relations. For example, in Hanoi-3x3, the obtained model has the single
action schema:

\begin{Verbatim}[frame=lines,label=hanoi-3x3,fontsize=\relsize{-3}]
MOVE(d,to,from):
 Static: NEQ(d,to), NEQ(d,from), NEQ(to,from), -BIGGER(d,to)
 Pre: -p(to,d), -p(from,from), p(d,d), p(to,to), p(from,d)
 Eff: -p(to,to), -p(from,d), p(to,d), p(from,from)
\end{Verbatim}

The predicate $p(x_1,x_2)$ denotes $on(x_2,x_1)$ if $x_1{\neq}x_2$, and $clear(x_1)$ otherwise.
The solver synthesizes   the \emph{exactly-1} invariant scheme $\{ p(o,x) : x \}$
for each object $o$ which  says that each disc must be clear or have another disc above it.

For Gripper-3, the solver learns the schemas:

\begin{Verbatim}[frame=lines,label=gripper-3,fontsize=\relsize{-3}]
MOVE(x,to,from)
  Static: NEQ(x,from), NEQ(to,from), -B1(x,to), -B2(x,x), B1(from,x)
  Pre: -Nat(from), Nat(x), Nat(to)
  Eff: -Nat(to), Nat(from)

DROP(g,b,r)
  Static: NEQ(g,r), -B2(r,b), B1(g,g)
  Pre: -Nat(r), -at-hold(r,b), Nfree(g), at-hold(g,b)
  Eff: -Nfree(g), -at-hold(g,b), at-hold(r,b)

PICK(g,r,b)
  Static: NEQ(g,r), -B2(r,b), B1(g,g)
  Pre: -Nat(r), -Nfree(g), -at-hold(g,b), at-hold(r,b)
  Eff: -at-hold(r,b), Nfree(g), at-hold(g,b)
\end{Verbatim}

Here, $at$-$hold(x,ball)$ represents $at(x,ball)$ if $x$ is a room,
and $hold(x,ball)$ if $x$ is a gripper.
% The \emph{exactly-1} invariant learned is $\{at$-$hold(x,ball)\}$
% that says that a ball can be either in a room or in a gripper.
% lack of space Hector
The binary static predicate $B_1(x_1,x_2)$ encodes that $x_1$ is a
gripper if $x_1=x_2$, or that $x_1$ and $x_2$ are rooms otherwise,
and $B_2(x_1, x_2)$ encodes pairs of objects $(x_1,x_2)$ that are
not a room and a ball.

The schemas learned for Blocks-2 are:

\begin{Verbatim}[frame=lines,label=blocks2-4,fontsize=\relsize{-3}]
NEWTOWER(x1,x2)
  Static: NEQ(x1,x2)
  Pre: -clear(x1), -p(x1,x1), clear(x2), p(x2,x1)
  Eff: -clear(x2), -p(x2,x1), p(x1,x1)

STACK(x1,x2)
  Static: NEQ(x1,x2)
  Pre: -clear(x1), -clear(x2), -p(x2,x1), p(x1,x1)
  Eff: -p(x1,x1), clear(x2), p(x2,x1)

MOVE(x1,x2,x3)
  Static: NEQ(x1,x2), NEQ(x1,x3), NEQ(x2,x3)
  Pre: -clear(x1), -clear(x3), -p(x3,x1), clear(x2), p(x2,x1)
  Eff: -clear(x2), -p(x2,x1), clear(x3), p(x3,x1)
\end{Verbatim}

In these schemas, the predicate $p(x_1,x_2)$ represents $on(x_2,x_1)$
or $ontable(x1)$ whether $x_1 \neq x_2$ or not. The solver finds the
invariant $\{p(x,o) : o\}$ that says that a block is either on top
of another, or on the table.

\section{Partial and Noisy Inputs}

A key assumption about the proposed learning paradigm,
borrowed from  Bonet and Geffner \shortcite{bonet:ecai2020},
is that the input graph is \textbf{complete} (no missing nodes or edges)
and \textbf{noise free} (no wrong edges and different nodes stand for
different states). We show next how to relax these assumptions.

\subsection{Partial Graphs}

The robustness to incomplete information is analyzed by feeding
the learner with a \textbf{partial graph} $G'$ of a true but hidden complete graph $G$.
The partial graph $G'$  contains some of the nodes $n$ from $G$,  some of the
edges among these  nodes from $G$, and counts of the number of outgoing
edges $(n,n')$ in $G$ that are not necessarily in $G'$, along with their labels.
% We construct the partial graph $G'$ with this information by performing random
% walks over $G$. Nodes and edges  traversed in the random walk are placed in $G'$,
% and when a node $n$ is visited, the number and labels of its outgoing edges are recorded.

The \textbf{learning task}  with a \textbf{partial graph} $G'$ is set to the following: find a
planning problem $P=\tup{D,I}$ and a mapping from the nodes $n$ in $G'$ to  unique states $s(n)$ over
$P$, and a mapping from the edges $(n,n')$ in $G'$ to ground actions $a(n,n')$ such that 1) action $a(n,n')$
is applicable in the state $s(n)$, 2)~the state $s(n')$ is the result of applying action $a(n,n')$ in the
state $s(n)$, and 3)~the set of actions in $P$ applicable in each state $s(n)$ match the outgoing edge
counts for $n$  (i.e., there are $k$ outgoing edges from $n$  with label $L$ iff
there are $k$ applicable actions in $s(n)$ with the same label). If the partial graph $G'$ is
the  complete graph $G$, the learning task is equivalent to the learning task
defined in Section~3.1.

In the experiments, we consider partial graphs $G'=G_p$ obtained by performing
a random walk in $G$ until $p$ per cent of the \emph{edges} in $G$ have been traversed.
The sampled graphs  $G_p$ represent information that an agent can acquire
in a real environment where different states can be distinguished.
\Omit{
  The agent starts in some random state, moves randomly, and observes states and state transitions.
  It is also reasonable to assume that, when in a state $s$, the agent %can
  also observes the \emph{number} of actions that can be taken in $s$ and their
  corresponding labels, but not necessarily the resulting states $s'$.
  This means that when a node $n$ is part of the sampled graph $G_p$, %we assume that
  the \emph{number of outgoing edges} $(n,n')$ and their
  labels are observable and available to the learner, although not necessarily
  the edges $(n,n')$ themselves.
}

The partial graphs $G_p$ are represented in the programs as the complete graphs $G$;
i.e., using the  atoms \atom|node(S)| and \atom|tlabel((S1,S2),L)| for  the nodes and the labeled edges in $G_p$, respectively.
In addition,  the representation of $G_p$  is extended with the atoms \atom|tlabel((S1,S2),L)| and \atom|unvisited(S1,S2)|
for every edge \atom|(S1,S2)| such that node \atom|S1| is visited but the edge \atom|(S1,S2)| is not.
Since, as we will see,  the program will make not use of the identity of the node \atom|S2| in these edges,
the atoms \atom|tlabel((S1,S2),L)| for unvisited edges provides just  a  way for encoding the outgoing node counts.
Indeed, the atoms \atom|tlabel((S1,S2),L)|  are used to activate the rules in lines \ref{asp:partial1:begin}--\ref{asp:partial1:end}
of the base code in Fig.~\ref{fig:program} and guarantee that the outgoing edges of \atom|S1|
have a corresponding ground action applicable, while the  \atom|unvisited(S1,S2)|  atoms are used to block the rules in lines
\ref{asp:partial2:begin}--\ref{asp:partial2:end} that relate the literals in the states associated with the nodes  \atom|S1| and \atom|S2|.
% In this way, information about the  nodes \atom|S2| in the atoms \atom|tlabel((S1,S2),L)| for untraversed edges is not used.
% so that such atoms actually end up encoding the node counts. This is accomplished by just adding the literal \atom|not\ unvisited(S1,S2)|
% to the rules about effects and inertia in lines \ref{asp:partial2:begin}--\ref{asp:partial2:end}
% of the base code. This  is the only difference between the program that works with sampled, partial graphs, and
% the program for complete graphs.
%
% In the end,
Every solution for the complete  graph $G$ is a solution of the partial graph  $G_p$ but not the other way around,
as $G_p$ may admit other models.

Experimental results about the performance of the learner over sampled
partial graphs $G_p$ are shown in Table~\ref{table:rollouts}.
For each planning instance $P$ considered that produced
a model that generalizes, we sample $10$ partial graphs $G_p$ from $G=G(P)$,
for each value of $p$ in $\{20,40,60,80\}$.
%For each $G_p$, the solver is given a maximum time window of 2 hours.
The table shows the percentage $p$ of edges sampled, the number of objects found, the number $n$ of
the 10 runs where a model was found, and the number of runs where the best model verified ($v$)
and was shown to be optimal ($opt$). Since the most common reason that partial graphs $G_p$
with a small number of sampled edges $p$  lead to action schemas that do not generalize
is that they are  too simple, we also report  the cost vector $V(M)$ for the best and worst
models found in the 10 runs. In these vectors  (Section~5.1), the three first  entries are the sum of action arities,
the sum of dynamic predicate arities, and the sum of static predicate arities.
It can be seen that in most domains, the models become more complex (costly) as the
number of sampled edges $p$ is increased. The exception is Blocks-2 where the ten models
found verify for almost all values of $p$, and they all have the same cost vectors.
% , given by
% action arities that add up to $7$, dynamic predicates of arities $2$ and $3$, and no static predicats.
% In particular,
The models that generalize in Blocks-2 are found for  $p=20$ ($20\%$ of the edges sampled).
In most domains, good  generalization results are  obtained for $p=60$.

\subsection{Noisy Edges and States}

We also tested  robustness by labeling certain edges $(n,n')$ in the input graph
as noisy, with the result that the states $s(n')$ represented by the nodes  $n'$ in such edges are \textbf{ambiguous}
and cannot be assumed to be different than the other  states. 
We can think of the nodes as represented by images,  for example, such  that it's not clear if a given image
is the same as an image already seen or not. It is up to the solver to resolve
these ambiguities. As before, the number of edges $(n,n')$ in the input graph that are considered
noisy is determined by  a sampling parameter $q$, that ranges between $1$ and $100$,
such that $q$ represents the percentage of edges that are \textbf{not} noisy, so that
$100-q$ is the percentage of noisy edges. In the experiments, the graph is assumed to be complete, but the formulation
can combine noisy edges and partial  graphs.

\Omit{%
  In this setting, whenever the agent applies an action and traverses a noisy edge,
  it cannot recognize if the next state has been previously visited. %is the same as some previously visited state. % or not.
  %
  %However, if the agent traverses a noisy edge $(n,n')$ more than once,
  %we assume that in all those transitions the agent recognizes that
  %the state associated with $n'$ is always the same.
  %
  As before, we assume that the agent observes the number and labels of the
  edges coming from visited states.
}

The graph $G_q$ that is fed to the solver
is the result of extending a partial or complete graph $G$
with  \emph{ambiguous} nodes that represent the states reached after traversing a noisy edge.
More in detail, the labeled edges $((n,n'),l)$
that are noisy are replaced by new labeled edges $((n,n^*),l)$ where $n^*$
is a new ambiguous node unique to the pair $(n,n')$.
To account for the uncertainty associated to them,
we allow their corresponding states to be the same as those
of other (ambiguous or non-ambiguous) nodes.
Next, for every labeled edge $((n',n''),l')$ in the resulting graph,
we add the new labeled edge $((n^*,n''),l')$.
This allows the ambiguous node $n^*$ to access the nodes that are reachable from $n'$,
some of which may be also ambiguous. The number of nodes in  $G_q$
can thus be larger than the number of nodes in $G$, with the excess nodes being
multiple  copies of the  noisy nodes.

The ASP encoding of this extension
introduces the facts \atom|ambiguous(n)|
for every ambiguous node \atom|n| in $G_q$, and
extends the constraint in line~\ref{asp:last:end} in Fig.~\ref{fig:program}
with the literals \atom|not\ ambiguous(S1)| and \atom|not\ ambiguous(S2)|
to allow ambiguous nodes to represent the same states as other nodes.
As with partial graphs, the solutions of the noisy graph $G_q$ are a superset of the solutions of the noise-free graph $G$.

Table~\ref{table:noise} shows results on complete graphs
for different percentages $q$ of \textbf{noise-free} edges.
%; the table contains columns for the value of $q$, the number of models that verify,
%and the average and standard deviation of time to find a best model for the noisy
%graph $G_q$ in the time window of 2 hours.
From the table, we observe that in regular and simple models like grid-v0 and
grid-v1, the approach is able to learn models that generalize even when there
are  many noisy edges. For more complex domains, as the parameter $q$ decreases
the models that verify tend to  decrease. The ambiguity resulting from the
``noise'' is not always detrimental to performance, however. In Gripper-2, for example,
for $q=40$, 9 out of the 10 runs  yield models that generalize,
while for $q=100$, only 6 of them do. In Gripper-3, with $q=80$,
4 out of the 10 runs yield models that generalize,
while for $q=100$, only 1 does.

\begin{table}[t]
  \centering
  \resizebox{\columnwidth}{!}{
    \begin{tabular}{@{}lrr@{\ \ \ }rrrrr@{}}
      \toprule
       %& & & & & \multicolumn{3}{c}{min} & \multicolumn{3}{c}{max} \\
       %\cmidrule{6-8} \cmidrule{9-11}
      Domain & \#obj & $p$ & $n$ & $v$ & $opt$ & $\min V(M)$ & $\max V(M)$ \\
      \midrule
      \begin{filecontents*}{results/statscost.csv}
instance|perc|solved|verify|optimum|labels|mina|minp|minpa|minpsa|minatom|maxa|maxp|maxpa|maxpsa|maxatom|inta|intp|intpa|intpsa|intatom|objs
arrow-5|10|10||10||2|1|1|0|2|3|1|1|2|2||||||
arrow-5|20|10||10||3|1|1|2|2|3|1|2|0|4||||||
arrow-5|40|10||10||3|1|2|1|4|3|1|2|1|4||||||
arrow-5|60|10||10||3|1|2|1|4|3|1|2|1|4||||||
arrow-5|80|10||10||3|1|2|1|4|3|1|2|1|4||||||
blocks1-4|10|8||2||6|3|3|1|8|12|3|5|2|8||||||
blocks1-4|20|3||0||10|3|4|3|5|12|3|4|4|10||||||
blocks1-4|40|0||0|||||||||||||||||
blocks1-4|60|0||0|||||||||||||||||
blocks1-4|80|0||0|||||||||||||||||
blocks2-4|20|10|10|3|3|7|2|3|0|7|7|2|3|0|7||||||4
blocks2-4|40|10|10|2|3|7|2|3|0|7|7|2|3|0|7||||||4
blocks2-4|60|10|9|1|3|7|2|3|0|7|7|2|3|0|8||||||4
blocks2-4|80|10|10|0|3|7|2|3|0|7|7|3|3|0|7||||||4
blocks2-4|100|10|10|0|3|7|2|3|0|7|7|3|3|0|7|7|2|3|0|8|4
grid-v0-3x4|10|10|0|10|4|3|3|1|2|3|7|1|1|2|2||||||4
grid-v0-3x4|20|10|0|10|4|6|2|2|2|4|8|1|1|4|3||||||4
grid-v0-3x4|40|10|6|10|4|7|2|2|2|3|8|2|2|2|2||||||4
grid-v0-3x4|60|10|10|10|4|8|2|2|2|2|8|2|2|2|2||||||4
grid-v0-3x4|80|10|10|10|4|8|2|2|2|2|8|2|2|2|2||||||4
grid-v0-3x4|100|10|10|10|4|8|2|2|2|2|8|2|2|2|2|8|2|2|4|2|4
grid-v1-3x4|10|10|0|10|2|2|2|1|2|1|4|1|1|2|3||||||4
grid-v1-3x4|20|10|2|10|2|4|1|1|2|2|4|2|2|2|2||||||4
grid-v1-3x4|40|10|5|10|2|4|2|2|2|2|4|2|2|2|2||||||4
grid-v1-3x4|60|10|9|10|2|4|2|2|2|2|4|2|2|2|2||||||4
grid-v1-3x4|80|10|5|10|2|4|2|2|2|2|4|2|2|2|2||||||4
grid-v1-3x4|100|10|9|10|2|4|2|2|2|2|4|2|2|2|2|4|2|2|4|2|4
gripper-2|10|10|0|10|3|5|1|1|2|3|6|2|2|2|4||||||4
gripper-2|20|10|0|9|3|5|3|3|2|3|7|2|2|2|4||||||4
gripper-2|40|10|1|0|3|6|3|4|3|5|8|3|3|3|5||||||4
gripper-2|60|10|4|0|3|8|1|2|1|6|9|3|5|4|7||||||4
gripper-2|80|10|8|0|3|8|2|3|2|5|8|3|5|4|10||||||4
gripper-2|100|10|6|0|3|8|3|4|1|6|9|3|5|3|10|8|2|3|4|5|4
gripper-3|20|2|1|0|3|8|2|3|4|7|9|3|5|4|10||||||4
gripper-3|40|1|1|0|3|9|3|5|4|10|9|3|5|4|10||||||4
gripper-3|60|1|0|0|3|9|3|4|3|10|9|3|4|3|10||||||4
gripper-3|80|3|2|0|3|8|3|5|3|7|9|3|5|3|10||||||4
gripper-3|100|1|1|0|3|9|3|5|3|9|9|3|5|3|9|8|2|3|4|6|4
hanoi-3x3|10|10|0|10|1|2|1|1|1|1|2|1|1|2|2||||||6
hanoi-3x3|20|10|0|5|1|2|1|1|2|2|3|1|1|2|2||||||6
hanoi-3x3|40|10|6|0|1|3|1|2|2|6|3|1|2|2|6||||||6
hanoi-3x3|60|10|10|0|1|3|1|2|2|6|3|1|2|2|6||||||6
hanoi-3x3|80|10|10|0|1|3|1|2|2|6|3|2|3|2|9||||||6
hanoi-3x3|100|9|9|0|1|3|1|2|2|6|3|2|3|2|6|3|1|2|2|6|6
\end{filecontents*}
\csvreader[separator=pipe,late after line=\\, filter=\equal{\domain}{blocks2-4}\and\not\equal{\perc}{10}]{results/statscost.csv}{%
        1=\domain,2=\perc,3=\solved,4=\verify,5=\opt,6=\act,7=\mina,8=\minp,9=\minpa,10=\minpsa,11=\minatom,12=\maxa,13=\maxp,14=\maxpa,15=\maxpsa,16=\maxatom,17=\ina,18=\inp,19=\inpa,20=\inpsa,21=\inatom,22=\objs}{%
        \domain & \objs & \perc & \solved & \verify & \opt & (\mina, \minp+\minpa, \minpsa) & (\maxa, \maxp+\maxpa, \maxpsa) }
      \midrule
      \csvreader[separator=pipe,late after line=\\, filter=\equal{\domain}{grid-v0-3x4}\and\not\equal{\perc}{10}]{results/statscost.csv}{%
      1=\domain,2=\perc,3=\solved,4=\verify,5=\opt,6=\act,7=\mina,8=\minp,9=\minpa,10=\minpsa,11=\minatom,12=\maxa,13=\maxp,14=\maxpa,15=\maxpsa,16=\maxatom,17=\ina,18=\inp,19=\inpa,20=\inpsa,21=\inatom,22=\objs}{%
        \domain & \objs & \perc & \solved & \verify & \opt & (\mina, \minp+\minpa, \minpsa) & (\maxa, \maxp+\maxpa, \maxpsa)  }
      \midrule
      \csvreader[separator=pipe,late after line=\\, filter=\equal{\domain}{grid-v1-3x4}\and\not\equal{\perc}{10}]{results/statscost.csv}{%
      1=\domain,2=\perc,3=\solved,4=\verify,5=\opt,6=\act,7=\mina,8=\minp,9=\minpa,10=\minpsa,11=\minatom,12=\maxa,13=\maxp,14=\maxpa,15=\maxpsa,16=\maxatom,17=\ina,18=\inp,19=\inpa,20=\inpsa,21=\inatom,22=\objs}{%
        \domain & \objs & \perc & \solved & \verify & \opt & (\mina, \minp+\minpa, \minpsa) & (\maxa, \maxp+\maxpa, \maxpsa)  }
      \midrule
      \csvreader[separator=pipe,late after line=\\, filter=\equal{\domain}{gripper-2}\and\not\equal{\perc}{10}]{results/statscost.csv}{%
      1=\domain,2=\perc,3=\solved,4=\verify,5=\opt,6=\act,7=\mina,8=\minp,9=\minpa,10=\minpsa,11=\minatom,12=\maxa,13=\maxp,14=\maxpa,15=\maxpsa,16=\maxatom,17=\ina,18=\inp,19=\inpa,20=\inpsa,21=\inatom,22=\objs}{%
        \domain & \objs & \perc & \solved & \verify & \opt & (\mina, \minp+\minpa, \minpsa) & (\maxa, \maxp+\maxpa, \maxpsa) }
      \midrule
      \csvreader[separator=pipe,late after line=\\, filter=\equal{\domain}{gripper-3}\and\not\equal{\perc}{10}]{results/statscost.csv}{%
      1=\domain,2=\perc,3=\solved,4=\verify,5=\opt,6=\act,7=\mina,8=\minp,9=\minpa,10=\minpsa,11=\minatom,12=\maxa,13=\maxp,14=\maxpa,15=\maxpsa,16=\maxatom,17=\ina,18=\inp,19=\inpa,20=\inpsa,21=\inatom,22=\objs}{%
        \domain & \objs & \perc & \solved & \verify & \opt & (\mina, \minp+\minpa, \minpsa) & (\maxa, \maxp+\maxpa, \maxpsa) }
      \midrule
      \csvreader[separator=pipe,late after line=\\, filter=\equal{\domain}{hanoi-3x3}\and\not\equal{\perc}{10}]{results/statscost.csv}{%
      1=\domain,2=\perc,3=\solved,4=\verify,5=\opt,6=\act,7=\mina,8=\minp,9=\minpa,10=\minpsa,11=\minatom,12=\maxa,13=\maxp,14=\maxpa,15=\maxpsa,16=\maxatom,17=\ina,18=\inp,19=\inpa,20=\inpsa,21=\inatom,22=\objs}{%
        \domain & \objs & \perc & \solved & \verify & \opt & (\mina, \minp+\minpa, \minpsa) & (\maxa, \maxp+\maxpa, \maxpsa) }
      \bottomrule
    \end{tabular}
  }
  %\vskip -.5em
  \caption{Learning from partial graphs $G_p$ where $p$ is the percentage of  sample edges from   $G(P)$ and  $P$ is the instance shown on the left. 
    Each  experiment is run 10 times, and  min and max  cost vectors  $V(M)$ of best models found are shown (Section~5). 
%     , indicating sum of action arities, sum of dynamic predicate
    %     arities (decomposed), and sum of static predicate arities (term  $N_g$ in cost vector omitted).
    % Hector commented
    Column $n$ shows how many solutions were found in the  10 runs, $v$ how many of them verify (generalize to larger instances),
    and $opt$ how many were proved optimal. For reference, the cost vectors of the  \textbf{intended models} for these domains
    are (7, 2+3, 0) for Blocks-2, (8,2+2,4) for Grid-v0, (4, 2+2, 4) for Grid-v1,  (8, 2+3, 4) for Gripper, and (3, 1+2, 2) for Hanoi.
     }
  \label{table:rollouts}
\end{table}

\begin{table}[t]
  \centering
  \resizebox{\columnwidth}{!}{
    \begin{tabular}{@{}lrr@{\ \ \ }rrrrr@{}}
      \toprule
       %& & & & & \multicolumn{3}{c}{min} & \multicolumn{3}{c}{max} \\
       %\cmidrule{6-8} \cmidrule{9-11}
      Domain & \#obj & $q$ & $n$ & $v$ & $opt$ & $\min V(M)$ & $\max V(M)$ \\
      \midrule
      \begin{filecontents*}{results/statscostnoise.csv}
instance|perc|solved|verify|optimum|labels|mina|minp|minpa|minpsa|minatom|maxa|maxp|maxpa|maxpsa|maxatom|inta|intp|intpa|intpsa|intatom|objs
blocks2-4|20|2|2|0|3|7|3|5|2|9|9|3|5|4|7||||||4
blocks2-4|40|6|6|0|3|7|3|5|0|7|9|3|5|4|7||||||4
blocks2-4|60|9|6|0|3|7|3|4|0|7|9|3|5|4|10||||||4
blocks2-4|80|9|9|0|3|7|2|3|0|7|8|3|5|0|8||||||4
blocks2-4|100|10|10|0|3|7|2|3|0|7|7|3|3|0|7|7|2|3|0|8|4
grid-v0-3x4|20|10|10|10|4|8|2|2|2|2|8|2|2|2|2||||||4
grid-v0-3x4|40|10|10|10|4|8|2|2|2|2|8|2|2|2|2||||||4
grid-v0-3x4|60|10|10|10|4|8|2|2|2|2|8|2|2|2|2||||||4
grid-v0-3x4|80|10|10|10|4|8|2|2|2|2|8|2|2|2|2||||||4
grid-v0-3x4|100|10|10|10|4|8|2|2|2|2|8|2|2|2|2|8|2|2|4|2|4
grid-v1-3x4|20|10|6|10|2|4|2|2|2|2|4|2|2|2|2||||||4
grid-v1-3x4|40|10|7|10|2|4|2|2|2|2|4|2|2|2|2||||||4
grid-v1-3x4|60|10|5|10|2|4|2|2|2|2|4|2|2|2|2||||||4
grid-v1-3x4|80|10|6|10|2|4|2|2|2|2|4|2|2|2|2||||||4
grid-v1-3x4|100|10|9|10|2|4|2|2|2|2|4|2|2|2|2|4|2|2|4|2|4
gripper-2|20|10|5|0|3|8|3|4|1|7|9|3|5|2|7||||||4
gripper-2|40|10|9|0|3|8|2|3|4|8|8|3|5|1|7||||||4
gripper-2|60|10|7|0|3|8|2|3|1|8|8|3|5|4|9||||||4
gripper-2|80|10|5|0|3|8|2|3|1|5|9|3|5|4|9||||||4
gripper-2|100|10|6|0|3|8|3|4|1|6|9|3|5|3|10|8|2|3|4|5|4
gripper-3|20|0|0|0|||||||||||||||||4
gripper-3|40|0|0|0|||||||||||||||||4
gripper-3|60|0|0|0|||||||||||||||||4
gripper-3|80|4|4|0|3|8|3|4|3|10|9|3|5|4|9||||||4
gripper-3|100|1|1|0|3|9|3|5|3|9|9|3|5|3|9|8|2|3|4|6|4
hanoi-3x3|20|0|0|0|1||||||||||||||||6
hanoi-3x3|40|1|1|0|1|3|1|2|2|9|3|1|2|2|9||||||6
hanoi-3x3|60|6|6|0|1|3|1|2|2|6|3|1|2|2|8||||||6
hanoi-3x3|80|4|4|0|1|3|1|2|2|6|3|2|3|2|7||||||6
hanoi-3x3|100|9|9|0|1|3|1|2|2|6|3|2|3|2|6|3|1|2|2|6|6
arrow_5|20|10||10||3|1|2|1|4|3|1|2|1|4||||||
arrow_5|40|10||10||3|1|2|1|4|3|1|2|1|4||||||
arrow_5|60|10||10||3|1|2|1|4|3|1|2|1|4||||||
arrow_5|80|10||7||3|1|2|1|4|3|1|2|1|4||||||
\end{filecontents*}
\csvreader[separator=pipe,late after line=\\, filter=\equal{\domain}{blocks2-4}\and\not\equal{\perc}{10}]{results/statscostnoise.csv}{%
        1=\domain,2=\perc,3=\solved,4=\verify,5=\opt,6=\act,7=\mina,8=\minp,9=\minpa,10=\minpsa,11=\minatom,12=\maxa,13=\maxp,14=\maxpa,15=\maxpsa,16=\maxatom,17=\ina,18=\inp,19=\inpa,20=\inpsa,21=\inatom,22=\objs}{%
        \domain & \objs & \perc & \solved & \verify & \opt & \ifcsvstrequal{\mina}{}{***}{(\mina, \minp+\minpa, \minpsa)} & \ifcsvstrequal{\mina}{}{***}{(\mina, \minp+\minpa, \minpsa)} }
      \midrule
      \csvreader[separator=pipe,late after line=\\, filter=\equal{\domain}{grid-v0-3x4}\and\not\equal{\perc}{10}]{results/statscostnoise.csv}{%
      1=\domain,2=\perc,3=\solved,4=\verify,5=\opt,6=\act,7=\mina,8=\minp,9=\minpa,10=\minpsa,11=\minatom,12=\maxa,13=\maxp,14=\maxpa,15=\maxpsa,16=\maxatom,17=\ina,18=\inp,19=\inpa,20=\inpsa,21=\inatom,22=\objs}{%
        \domain & \objs & \perc & \solved & \verify & \opt & \ifcsvstrequal{\mina}{}{***}{(\mina, \minp+\minpa, \minpsa)} & \ifcsvstrequal{\mina}{}{***}{(\mina, \minp+\minpa, \minpsa)} }
      \midrule
      \csvreader[separator=pipe,late after line=\\, filter=\equal{\domain}{grid-v1-3x4}\and\not\equal{\perc}{10}]{results/statscostnoise.csv}{%
      1=\domain,2=\perc,3=\solved,4=\verify,5=\opt,6=\act,7=\mina,8=\minp,9=\minpa,10=\minpsa,11=\minatom,12=\maxa,13=\maxp,14=\maxpa,15=\maxpsa,16=\maxatom,17=\ina,18=\inp,19=\inpa,20=\inpsa,21=\inatom,22=\objs}{%
        \domain & \objs & \perc & \solved & \verify & \opt & \ifcsvstrequal{\mina}{}{***}{(\mina, \minp+\minpa, \minpsa)} & \ifcsvstrequal{\mina}{}{***}{(\mina, \minp+\minpa, \minpsa)} }
      \midrule
      \csvreader[separator=pipe,late after line=\\, filter=\equal{\domain}{gripper-2}\and\not\equal{\perc}{10}]{results/statscostnoise.csv}{%
      1=\domain,2=\perc,3=\solved,4=\verify,5=\opt,6=\act,7=\mina,8=\minp,9=\minpa,10=\minpsa,11=\minatom,12=\maxa,13=\maxp,14=\maxpa,15=\maxpsa,16=\maxatom,17=\ina,18=\inp,19=\inpa,20=\inpsa,21=\inatom,22=\objs}{%
        \domain & \objs & \perc & \solved & \verify & \opt & \ifcsvstrequal{\mina}{}{***}{(\mina, \minp+\minpa, \minpsa)} & \ifcsvstrequal{\mina}{}{***}{(\mina, \minp+\minpa, \minpsa)} }
      \midrule
      \csvreader[separator=pipe,late after line=\\, filter=\equal{\domain}{gripper-3}\and\not\equal{\perc}{10}]{results/statscostnoise.csv}{%
      1=\domain,2=\perc,3=\solved,4=\verify,5=\opt,6=\act,7=\mina,8=\minp,9=\minpa,10=\minpsa,11=\minatom,12=\maxa,13=\maxp,14=\maxpa,15=\maxpsa,16=\maxatom,17=\ina,18=\inp,19=\inpa,20=\inpsa,21=\inatom,22=\objs}{%
        \domain & \objs & \perc & \solved & \verify & \opt & \ifcsvstrequal{\mina}{}{***}{(\mina, \minp+\minpa, \minpsa)} & \ifcsvstrequal{\mina}{}{***}{(\mina, \minp+\minpa, \minpsa)} }
      \midrule
      \csvreader[separator=pipe,late after line=\\, filter=\equal{\domain}{hanoi-3x3}\and\not\equal{\perc}{10}]{results/statscostnoise.csv}{%
      1=\domain,2=\perc,3=\solved,4=\verify,5=\opt,6=\act,7=\mina,8=\minp,9=\minpa,10=\minpsa,11=\minatom,12=\maxa,13=\maxp,14=\maxpa,15=\maxpsa,16=\maxatom,17=\ina,18=\inp,19=\inpa,20=\inpsa,21=\inatom,22=\objs}{%
        \domain & \objs & \perc & \solved & \verify & \opt & \ifcsvstrequal{\mina}{}{***}{(\mina, \minp+\minpa, \minpsa)} & \ifcsvstrequal{\mina}{}{***}{(\mina, \minp+\minpa, \minpsa)} }
      \bottomrule
    \end{tabular}
  }
  %\vskip -.5em
  \caption{Learning with noisy edges and states.  Results shown  on complete graphs with a percentage $q$ of edges that are not noisy  over 10 runs, along with
    minimum and maximum cost vectors  $V(M)$ of best models found.   The column $n$ shows how many solutions found in  these 10 runs, $v$ how many of them verify,
    and $opt$ how many were proved optimal. Asterisks indicate that  no model was  found in the 10 runs.
  }
  \label{table:noise}
\end{table}

\section{Conclusion}

We have explored variations and extensions of the approach to representation learning
for planning proposed by \citeay{bonet:ecai2020} that showed
how crisp and meaningful symbolic representations can be learned
from flat state spaces associated with small instances, in a way that
the resulting first-order domains (action schemas and predicates)
generalize to larger instances. The new ASP implementation has been shown to be
more scalable and robust, as it opens new possibilities for modeling and solving the
learning problem, and a higher level of abstraction to explore them.
The performance improvements are  significant, as the consideration
of thousands of propositional SAT theories encoding all possible values of the hyperparameters,
have been replaced by a simple meta-search on the number of objects, as all other
decisions are left to the solver, which in many cases manages to find solutions
and prove them optimal (simplest). In the new, high level ASP encoding,
it has also been simple to relax some of the assumptions made in previous work, namely
that the input graphs are complete and noise-free.
% Indeed, the new encodings
% handle naturally partial information about the state graphs, as well as noise
% that prevents some states from being distinguished from others.
It remains as an interesting challenge to improve the experimental results
even further so that the resulting methods can be applied to
learn representations, for example, % from labeled graphs, for example,
from arbitrary IPC planning domains.
% The proposed encodings provide
% a good basis from which this challenge can be pursued.

\section*{Acknowledgements}

The work is partially supported by an ERC Advanced Grant (No 885107),
by project TAILOR, funded by an EU Horizon 2020 Grant (No 952215),
and by the Knut and Alice Wallenberg (KAW) Foundation under the WASP
program. Hector Geffner is a Wallenberg Guest Professor at Link\"oping
University, Sweden.

% \vfill
% \pagebreak

\small

%% The file kr.bst is a bibliography style file for BibTeX 0.99c
\bibliographystyle{kr}
\bibliography{control}

\end{document}